\documentclass[runningheads]{llncs}
\usepackage{graphicx}

\usepackage{tikz}
\usepackage{comment}
\usepackage{amsmath,amssymb}
\usepackage{color}

\usepackage[pagebackref,breaklinks,colorlinks]{hyperref}
\usepackage{multibib}
\newcites{appendix}{\;}

\usepackage[accsupp]{axessibility} 
\usepackage[width=122mm,left=12mm,paperwidth=146mm,height=193mm,top=12mm,paperheight=217mm]{geometry}

\usepackage{xspace}
\makeatletter
\DeclareRobustCommand\onedot{\futurelet\@let@token\@onedot}
\def\@onedot{\ifx\@let@token.\else.\null\fi\xspace}
\def\eg{\emph{e.g}\onedot} 
\def\ie{\emph{i.e}\onedot}

\def\etal{\emph{et al}\onedot}
\makeatother

\usepackage{algorithm}
\usepackage[noend]{algpseudocode}
\algnewcommand\algorithmicinput{\textbf{Input: }}
\algnewcommand\Input{\item[\algorithmicinput]}
\usepackage{enumitem}
\usepackage{graphicx}
\usepackage{booktabs}
\usepackage{pifont}
\usepackage{multirow}
\usepackage{tabularx}
\newcolumntype{Y}{>{\centering\arraybackslash}X}
\usepackage{wrapfig}

\newcommand{\cmark}{\ding{51}}%
\newcommand{\xmark}{\ding{55}}%

\usepackage[labelformat=simple]{subcaption}
\begin{document}
\pagestyle{headings}
\mainmatter

\title{Self-supervised HDR Imaging \\ from Motion and Exposure Cues} 

\titlerunning{Self-supervised HDR from Motion and Exposure Cues}

\author{Michal Nazarczuk \qquad
Sibi Catley-Chandar  \\
Ales Leonardis \qquad
Eduardo Pérez Pellitero}

\authorrunning{M. Nazarczuk et al.}
\institute{Huawei Noah's Ark Lab}

\maketitle

\begin{abstract}
     Recent High Dynamic Range (HDR) techniques extend the capabilities of current cameras where scenes with a wide range of illumination can not be accurately captured with a single low-dynamic-range (LDR) image. This is generally accomplished by capturing several LDR images with varying exposure values whose information is then incorporated into a merged HDR image. While such approaches work well for static scenes, dynamic scenes pose several challenges, mostly related to the difficulty of finding reliable pixel correspondences. Data-driven approaches tackle the problem by learning an end-to-end mapping with paired LDR-HDR training data, but in practice generating such HDR ground-truth labels for dynamic scenes is time-consuming and requires complex procedures that assume control of certain dynamic elements of the scene (\eg~actor pose) and repeatable lighting conditions (stop-motion capturing). In this work, we propose a novel self-supervised approach for learnable HDR estimation that alleviates the need for HDR ground-truth labels. We propose to leverage the internal statistics of LDR images to create HDR pseudo-labels. We separately exploit static and well-exposed parts of the input images, which in conjunction with synthetic illumination clipping and motion augmentation provide high quality training examples. Experimental results show that the HDR models trained using our proposed self-supervision approach achieve performance competitive with those trained under full supervision, and are to a large extent superior to previous methods that equally do not require any supervision. 
\end{abstract}


\section{Introduction}
\label{sec:intro}
\begin{figure}[t]
    \centering
    \includegraphics[width=0.7\linewidth]{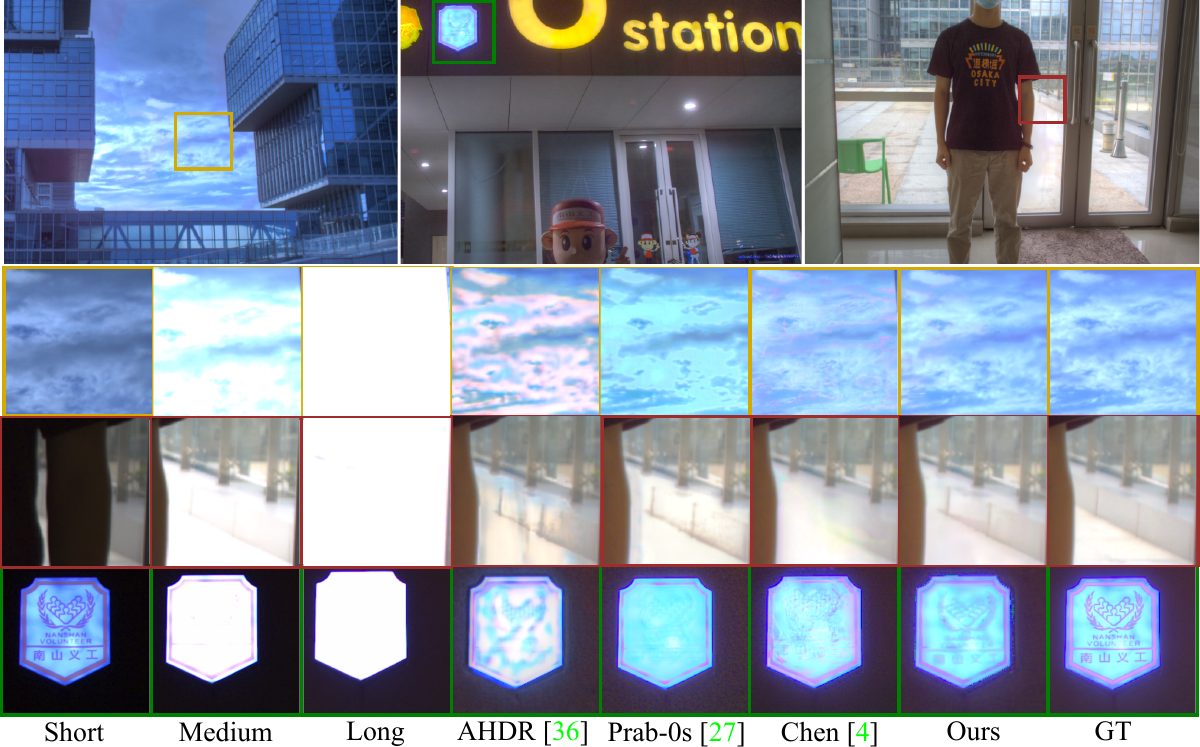}
    \caption{A qualitative comparison of our proposed self-supervised approach against other supervised \cite{Yan19,Chen21} and weakly-supervised \cite{Prabhakar21} methods. We introduce an HDR training strategy that does not require any ground-truth and performs quantitatively better than other unsupervised algorithms and comparable to supervised ones (Tab.~\ref{tab:results} and \ref{tab:results_chen}).}
    \label{fig:teaser}
    \vspace{-2em}
\end{figure}

While most uniformly illuminated scenes can be perfectly captured with a conventional camera, it is not uncommon to encounter scenes where the underlying dynamic range, \ie the ratio between the maximum and minimum irradiance, is such that said conventional cameras can not properly capture highlights and shadows simultaneously, and thus suffer from under- and over-exposed pixels. HDR imaging techniques fill in the gap and aim at reconstructing accurate image representations for scenes whose dynamic range is typically beyond three orders of magnitude~\cite{Mann16}.

The principle behind most HDR capturing strategies rely on obtaining different exposures of the scene that are then incorporated into a single HDR reconstruction \cite{Debeveck97}. This can be achieved by multi-camera systems capturing synchronously the scene \cite{McGuire07,Tocci11,Froehlich14} but arguably those have been so far less appealing due to \eg\,increased cost and fragility. Techniques that utilise a single sensor and merge frames captured at different time instants \cite{Debeveck97} have been proven more pragmatic, and are nowadays widely adopted in consumer cameras of all budgets.

Early frame fusion methods work well under static scenes \cite{Debeveck97,Mertens07}, however degrade quickly when there is complex motion (\eg~ghosting artefacts). Even though some methods proposed the use of \textit{off-the-shelf} alignment algorithms to find correspondences across frames \cite{Ward03}, it is the recent advances of deep learning methods that have established a new \textit{state-of-the-art} by learning an end-to-end mapping between the unaligned LDR input images and the target HDR domain \cite{Wu18,Yan19}. For training such supervised methods, training data where input dynamic scenes are paired to HDR ground-truth images is required.

The seminal work of Kalantari and Ramamoorthi \cite{Kalantari17} propose for the first time a data-capturing protocol to collect paired dynamic scenes and respective ground-truth labels, which has been adopted in recent datasets \cite{Chen21,Prabhakar19}. Firstly, a subject is asked to stay still and three bracketed exposure images are obtained on a tripod (\textit{static set}) which are then combined to produce the ground truth image. Later, the subject is asked to move and another set of bracketed exposure images is captured (\textit{dynamic set}). The input set is formed by taking the low and high exposure images from this dynamic set and the middle exposure image from  the static set. This procedure however comes with substantial drawbacks: it is complex and time-consuming, and most importantly, it assumes control over the dynamic elements of the scene (\eg~actor pose) and the repeatability of the scene (\eg~composition, illumination) which in essence limit the diversity of scenes and motions that can be captured.

In this work we propose a novel self-supervised HDR methodology that enables training deep models without the need of any ground-truth HDR image. To the best of our knowledge, no other HDR learning-based previous work has proposed this set-up. In our work, we build on the key observation that LDR input images contain useful information that is transferable to the HDR estimation task when certain conditions are met. In other words, we can systematically study and disentangle the factors of degradation from the HDR to LDR domains, and choose accordingly regions of the LDR images that are not degraded and can thus serve as valuable supervision. 

For that purpose, we consider two domains of supervision: (a) the Motion Domain and (b) the Exposure Domain which we then use to generate paired HDR pseudo-labels. The intuition behind (a) is that presence of motion can be automatically and locally determined (\ie as opposed to an image-level rigid \textit{dynamic vs static} label) and therefore pseudo-labels flexibly obtained; and behind (b) that well-exposed regions in the input LDR images are good local approximations of the HDR image, and thus can be used directly as HDR pseudo-labels. These two criteria, in conjunction with further simple synthetic-illumination modelling and motion augmentation enable effective and balanced supervision for challenging dynamic HDR benchmarks \cite{Kalantari17,Chen21}.

In summary, the contributions of this paper are: \textbf{(1)} A novel strategy to create HDR pseudo-labels from LDR images based on motion and exposure characteristics, \textbf{(2)} a  mechanism to transfer and synthetize over-exposed patches via gain mask for well-exposed patches and \textbf{(3)} comprehensive experiments and ablation studies to demonstrate the effectiveness of our proposed approach. 

\section{Related work}
In this section we provide an overview of relevant multiframe HDR methods that use Convolutional Neural Networks (CNN) and discussion about weakly- and self-supervised approaches. For a more complete review of the HDR SOTA we refer the reader to \cite{Wang21}.

\textbf{HDR Methods}: Together with their paired dataset, Kalantari and Ramamoorthi proposed an HDR fusion method composed by two stages: alignment and fusion. Firstly, input frames are aligned via optical flow \cite{Liu09} and then a CNN is used to merge aligned frames. Wu \etal~\cite{Wu18} adopted a similar scheme, however preferring a \textit{simple} global homography rather than a dense optical flow field for the alignment step, and a UNet-like architecture for the fusion of images. These two methods rely on the CNN to suppress errors on alignment at the fusion stage, but do not have an explicit mechanism within their architectures to regulate the contribution of each input frame. Yan \etal~\cite{Yan19} introduce an attention mechanism that is able to select or suppress features from each respective input frame, which improves performance both for unaligned or pre-aligned input frames. Shortly after, Yan \etal~\cite{Yan20} explore the non-local correlation in inputs frames in order to reduce ghosting artefacts. In the work of Prabhakar \etal~\cite{Prabhakar20} parts of the computation, including the optical flow estimation, are performed in a lower resolution and later upscaled back to full resolution using a guide image generated with a simple weight map, thus saving some computation. Recently, Chen \etal~\cite{Chen21} estimate HDR video streams, and provide a new test set and related benchmark that retains similarities with \cite{Kalantari17} in the capturing procedure.

\begin{figure*}[t]
    \centering
    \includegraphics[width=0.93\textwidth, trim=0 0 30 0, clip]{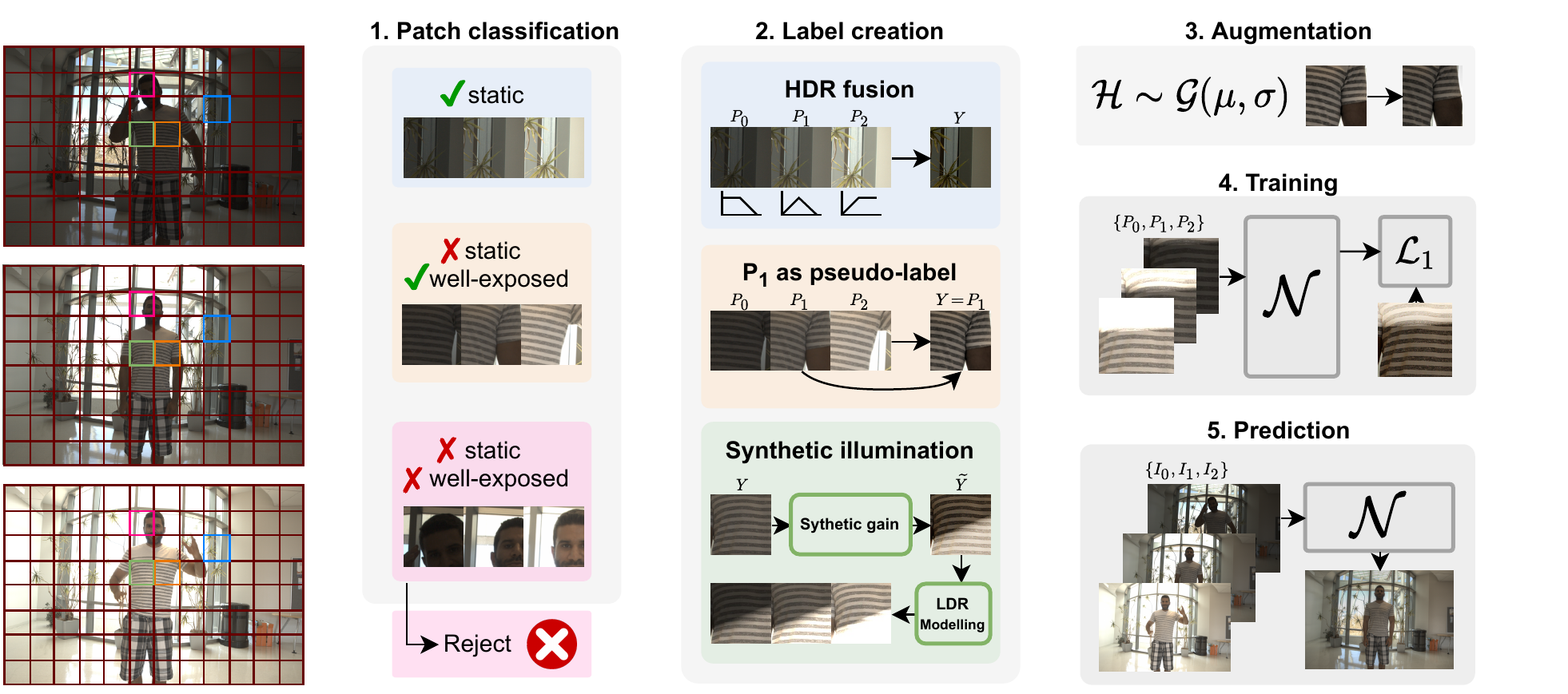}
    \caption{An overview of our method. Input LDR images are considered on a patch level. Image patches are processed to assess their motion and exposure characteristics. HDR fusion is performed for patches that after alignment have no motion. Additionally, we search for dynamic patches, for which the reference frame can serve as an HDR estimation. All well-exposed patches that can be an HDR pseudo-label are applied with synthetic illumination gain, from which an LDR reconstruction is performed. All obtained supervision pairs along with their counterparts augmented with synthetic camera movement are provided as supervision pairs for HDR reconstruction training.}
    \label{fig:method_overview}
    \vspace{-2em}
\end{figure*}

\textbf{Self-supervised Learning}: All the discussed methods above require training with full ground-truth supervision. As discussed in Section~\ref{sec:intro}, capturing HDR labels is inconvenient, time-consuming and inherently limits the scene and motion variability. Self-supervised learning (SSL) has attracted much interest during the last years in part for its clear success on representation learning \cite{Chen20,Asano20,He20,Longlong21}, but also due to its general appeal for problems where capturing large amount of data is still an open problem. Literature has mainly focused on high-level tasks, however there have been a number of works that explore self-supervision for \eg~image denoising \cite{Lehtinen18,Laine19}, super resolution~\cite{Wang21}. These works however are not directly applicable to the HDR estimation task, as some of the necessary assumptions on the estimated residual distribution do not hold, or still require some form of supervision. The recent work of Prabhakar \etal~\cite{Prabhakar21} is the first attempt at training-data efficiency applicable to HDR imaging. They propose a novel weakly-supervised training strategy, where they use \textit{easy-to-capture} static scenes in combination with unlabelled dynamic sequences and only a few \textit{costly} dynamic HDR labels, and achieve results that are competitive to fully supervised methods. Their strategy is divided into two stages: Firstly they train their model with the few labelled dynamic scenes, and use that model to predict HDR pseudo-labels from the pool of unlabelled dynamic data. In stage two the model is further fine-tuned with all the available original data and the generated pseudo-labels. Despite the important contributions of this work, the method does still require some manual labelling effort, \ie~whole image static scenes need to be captured with a tripod, and ideally a number of dynamic scenes captured in the stop-motion fashion.

\section{Method} \label{sec:method}

Existing data-driven approaches for HDR reconstruction use either full, or weak supervision from ground truth HDR images. Those HDR labels are 
very 
difficult to capture and require a specific setup (static camera and controllable elements of the scene, \eg moving actor). We propose a method that alleviates the need for ground truth labels and is applicable to various learning-based approaches for HDR reconstruction. We introduce a self-supervision strategy that leverages various LDR image characteristics
to produce supervision signals that do not require access to 
ground truth HDR image. Hence, this allows for the use of any varying exposure LDR sequence captured with no constraints on the procedure. 

Given the set of LDR images with varying exposure $(I_0, I_1, I_2)$, where $I_1$ refers to an arbitrarily chosen reference image, our goal is to produce a set of HDR pseudo-labels in the form of sets of patches  
$\{(P_{0i}, P_{1i}, P_{2i}), Y_i\}$
suitable for use as a supervision signal in data-driven approaches. 

We observe that there are two main conditions for a straightforward capture and reconstruction of HDR patches: (1) the scene, or at least a part of it, is static and allows thus to reliably perform a linear combination of input LDR images with varying exposure; or (2) the LDR reference image is well-exposed on its own, \ie does not contain over- or under-exposed regions or any form of degradation. We argue that sequences of LDR images often contain regions that satisfy one of the given conditions, and by using patches rather than images we can flexibly select them. Static parts of the image can be used to reconstruct HDR via direct linear fusion. Well-exposed image regions represent areas where input images provide a good approximation of the underlying luminosity and can thus be used as HDR pseudo-labels. We introduce further synthetic illumination changes on those pseudo-labels such that the respective LDR patches contain information loss due to over-exposure, and thus create valuable LDR to HDR supervision on dynamic sets with over-exposed regions.
In summary, we propose a method that exploits static and well-exposed regions in the image separately to produce HDR image patches for self-supervision. Figure \ref{fig:method_overview} presents an overview of the proposed method. A set of input images is divided into square patches $\{P_{0i}, P_{1i}, P_{2i}\}$ and parsed 
by characteristics classification and label creation modules.
Sets of patches and created pseudo-labels provide supervision LDR-HDR pairs that can be used by any learning method. Additionally, we propose the use of simple motion augmentation on all patches to provide additional supervision for dynamic, misaligned patches.

\subsection{Motion domain}

In the proposed approach, we suggest to obtain a set of pseudo-labels for HDR supervision based on motion characteristics of the image. The intention is to extract image regions suitable to undergo static HDR fusion.
We perform classification based on the optical flow estimation of the unlabelled input data. A high level overview of the the method is shown in Algorithm \ref{alg:static}. Firstly we estimate optical flow between each image and the reference image, in both directions (\ie $I_1I_0$, $I_0I_1$, $I_1I_2$, $I_2I_1$). We use the recent GMA~\cite{Jiang21} pretrained on the Sintel dataset \cite{Butler12}. In order to provide an HDR estimation, 
images have to be fully aligned.
We assume that very small camera motions can be easily corrected. 
We create a histogram of optical flow magnitudes and consider only images with dominant optical flow magnitude below a given threshold $t_{f}$. 
The camera movement is corrected by estimating a homography transformation $H_{i1}$ between all images with respect to the reference frame. 
Further, we apply warping to all non-reference images ($I_0$, $I_2$) to the reference image coordinate system obtaining thus an aligned set of images $(I_{0W}, I_1, I_{2W})$. For optical flow visualisations and details on alignment see Supplementary Material.

\vspace{-2em}
\begin{algorithm}
\caption{An algorithm for patch classification.}\label{alg:static}
    \begin{algorithmic}
        \Input LDR: $(I_0, I_1, I_2)$
        \State $\{P_{0i}, P_{1i}, P_{2i}\} \gets patches(I_0, I_1, I_2)$ 
        \State \textit{optical flow ($OF$)} $\gets f(I_j, I_k)$
        \If{$mode(OF\{I_1, I_2, I_3\}) < t_f$}
            \State $H_{01}, H_{21} \gets \mathtt{RANSAC}(I_0, I_1), \mathtt{RANSAC}(I_2, I_1)$ 
            \State $I_{0W}, I_{2W} \gets warp(I_0, H_{01}), warp(I_2, H_{21})$ 
            \State $\{P_{0Wi}, P_{2Wi}\} \gets patches(I_{0W}, I_{2W})$
            \For{$i=0...N$}
                \If{$(P_{0i}, P_{1i}, P_{2i})$ \textit{static}}
                \State $Y_{i} \gets \Lambda_0 P_{0Wi} + \Lambda_1 P_{1i} + \Lambda_2 P_{2Wi} $ 
                \State \textit{Save: } $((P_{0i}, P_{1i}, P_{2i}),\ Y_{i})$
                \Else
                    \If{$P_{1i}$ \textit{well-exposed}} 
                    \State $Y_{i} \gets P_{1i} $ 
                    \State \textit{Save: } $((P_{0i}, P_{1i}, P_{2i}),\ Y_{i})$
                    \EndIf
                \EndIf
            \EndFor
        \Else
            \For{$i=0...N$}
                \If{$P_{1i}$ \textit{well-exposed}} 
                \State $Y_{i} \gets P_{1i} $ 
                \State \textit{Save: } $((P_{0i}, P_{1i}, P_{2i}),\ Y_{i})$
                \EndIf
            \EndFor
        \EndIf
    \end{algorithmic}
\end{algorithm}
\vspace{-2em}

Thereafter, we extract patches from the warped images $\{P_{0Wi}, P_{2Wi}\}$ corresponding directly to patches from original images. Further we consider each set of patches separately and classify them as \textit{static} or \textit{dynamic}. We measure the difference of the optical flow magnitude with respect to its median and compare to a given threshold (we set a threshold $t_s$ as a function of the median $m$: $T=max(min(m, 2), 0.5)$ to allow for slightly more lenience with bigger movements). The condition has to be satisfied by all the computed optical flows across all aforementioned pairs of images to consider the patch as static.

If the patch is considered \textit{static}, we perform HDR fusion based on the warped set of patches $(P_{0Wi}, P_{1i}, P_{2Wi})$. Merging is done by computing a weighted combination of patches. We use a triangular weighting scheme similar to Debevec \etal \cite{Debeveck97} and Kalantari \etal \cite{Kalantari17}. The fusion is done in the linear image domain and we assume a gamma shaped curve for conversion. Finally, we perform a last consistency check to ensure the HDR reconstruction has gone well: we reject static ground truth patches that have a low PSNRs when compared to the well-exposed regions of the input LDR image (see Subsection \ref{subsec:well-exposed}) in order to avoid misalignment and ghosting artefacts in the pseudo-labels. The resulting HDR patches alongside the respective input LDR patches constitute LDR-HDR supervision pair $((P_{0i}, P_{1i}, P_{2i}),\ Y_{i})$. 

\begin{figure}
    \vspace{-1em}
    \centering
    \includegraphics[width=0.96\linewidth, trim=0 0 40 0, clip]{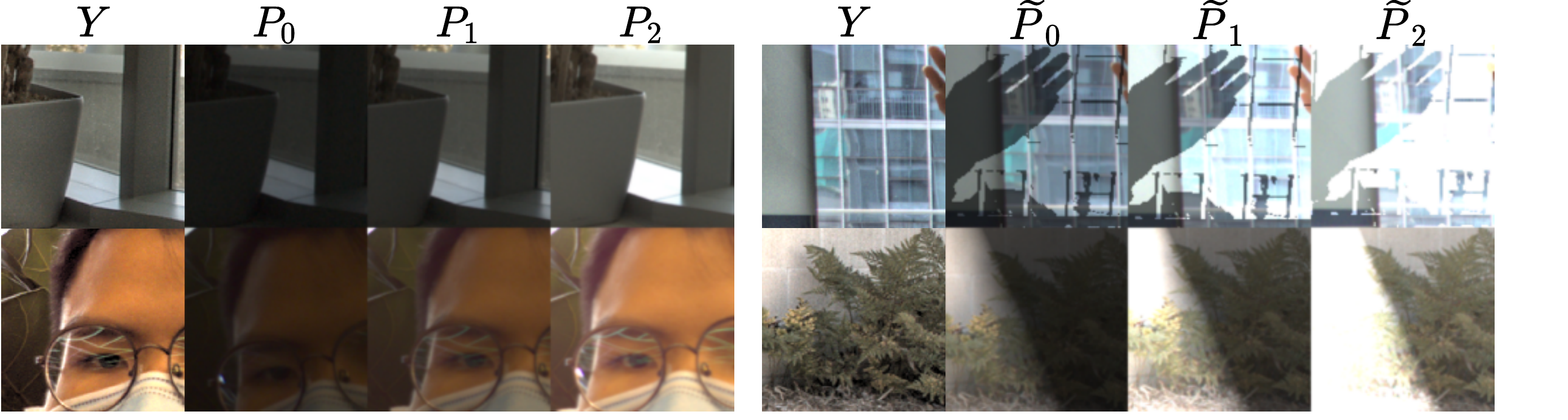}
    \caption{Example of HDR pseudo-label with short, medium, long LDR patches. \textbf{Left:} HDR pseudo-label from \textit{static} patch (top), and well-exposed reference patch used as pseudo-label (bottom). \textbf{Right:} Original pseudo-label and patches obtained with synthetic illumination and LDR modelling for transfer (top) and synthetic (bottom) masks.}
    \label{fig:example}
    \vspace{-3em}
\end{figure}

For all \textit{dynamic} patches (including patches from highly misaligned images), 
we check if the reference frame is well-exposed (see Subsection \ref{subsec:well-exposed}). If the patch is considered reliably exposed, we use the reference patch directly as the HDR pseudo-label and create a supervision pair $((P_{0i}, P_{1i}, P_{2i}),\ Y_{i})$ with real dynamic motion. These examples provide useful real dynamic training pairs that can guide methods on how to align LDR inputs to the HDR reference frame. We show in Figure \ref{fig:example} (left) examples of our proposed HDR pseudo-labels alongside corresponding LDR patches for both static and dynamic stacks of input patches.

\subsection{Exposure domain} \label{subsec:well-exposed}
Intuitively, HDR static fusion weights \cite{Debeveck97,Kalantari17} provide higher confidence in areas of the image that are well-exposed given each exposure value. Similarly, we locally select regions of the LDR images that are well exposed and use those images as HDR pseudo-labels. We then synthetically extend their dynamic range so that they provide valuable supervision on the exposure domain.
Firstly, for each 
patch we assess whether it is well-exposed. We perform that 
via
comparison to lower and upper threshold of the fusion triangular functions (\ie pixel illumination values) \cite{Kalantari17} and consider a patch well-exposed if the majority of pixels lay within the given range. In such a way we reject patches containing over- and under-exposed regions. We consider such well-exposed patches as a reliable estimation for the corresponding HDR patch. 
Further, we apply a synthetic illumination change to the patch, providing 
an HDR pseudo-label whose range of values goes beyond that of the original LDR patch, \ie it has a higher range of illumination. We apply this illumination change as a gain mask over the patch. We consider two alternatives for creating the mask:
\begin{enumerate}[itemsep=1pt,topsep=2pt,parsep=0pt]
\item \textbf{transfer} -- saturation masks obtained from non-well-exposed patches (selected randomly),
\item \textbf{synthetic} -- randomly generating a line going through the patch and applying mask increasing progressively towards the patch boundary (resembling the sun falling onto the planar surface).
\end{enumerate}
The value of the gain applied for the masked pixels is adjusted such that a given percentage of pixels within the patch become saturated. 
These synthetically generated HDR pseudo-labels are re-exposed (and clipped) according to the exposure values of the original LDR input following an image formation model \cite{Hasinoff10,Perez21}. Examples of original well-exposed patches with corresponding sets of generated LDR patches are presented in Figure \ref{fig:example} (right). For more examples of generated supervision pairs (both motion and exposure domains) we refer the reader to the Supplementary Material.

\subsection{Movement augmentation}
To complement and further balance the dynamic and static pseudo-labels we introduce simple motion augmentation. We consider two modes of handheld photo shooting:  \textit{pseudo-static}, similar to \eg~the camera shake induced by a person taking a picture while trying to keep the camera steady; and \textit{free-moving} when the camera is held freely with no consideration of the movement. We introduce global camera motion as a form of augmenting the generated supervision pairs. For the movement generation we randomly choose values of horizontal and vertical displacement from a Gaussian distribution (mean $0$ and given standard deviation for pseudo-static movement, higher mean and 
deviation for the \textit{free-moving} mode). Pseudo-static movement is applied only to synthetically generated supervision pairs from well-exposed patches as it is assumed to be naturally present in the 
pairs estimated from static patches. Due to the nature of our approach, most of the high-movement patches are filtered out, and thus we apply larger displacement values to both sets of generated supervision pairs.

\subsection{Model} 

Our self-supervision method provides a supervision of LDR image sets paired with respective HDR pseudo-label. Therefore, it can be applied to any existing data-driven approach, \eg any HDR estimation network. In our experiments we have chosen to focus on a simple UNet-based architecture\cite{Ronneberger15}, as it is a  well-explored backbone, and has proven effective in several image-to-image translation problems \cite{Badrinarayanan17,Isola17,Liu18}. Our selected UNet architecture contains 4 down- and upsampling modules and is used to predict an HDR residual signal from the stack of channel-wise concatenated inputs. Additionally, a reference input with over- and underexposed regions masked out is passed though a shallow convolutional module to produce a vanilla attention map which is later used to guide the merging of reference image and residual signal. 
We use a weighted sum of mean absolute errors for linear and $\mu$-law tonemapped images ($\mathcal{T}(I)=\frac{log(1+\mu I)}{log(1 + \mu)}$ where $\mu=5000$) as a loss function (Equation \ref{eq:loss}). More details on the network architecture are provided in the Supplementary Material.
\begin{equation}
\vspace{-1em}
\label{eq:loss}
    \textit{l}(I, GT) = \mathcal{L}_1(I, GT) + \alpha \mathcal{L}_1(\mathcal{T}(I), \mathcal{T}(GT))
\end{equation}


\section{Experiments}

We test our proposed approach on the Kalantari \etal dataset \cite{Kalantari17} and the recent datasets introduced and used by Chen \etal~\cite{Chen21}: \textbf{D} - dynamic dataset with ground truth, \textbf{DnGT} - dynamic dataset without ground truth, \textbf{S} - static dataset, \textbf{SRM} - static dataset with random synthetic movement, \textbf{HdM2} - 2 sequences from HdM-HDR \cite{Froehlich14}.
We consider video sequences as separate triplets of images with alternating exposures. 

All sets of LDR images were processed as described in Section \ref{sec:method}. We used patches of size $(128, 128)$, extracted with stride $64$. The threshold for the well-exposed values was set to $T_L=0.125$, $T_H=0.75$ for under- and overexposure respectively (for images considered in sRGB domain). Additionally, saturation masks from non-well-exposed patches were used only if covered at least $10\%$ of the patch surface. In the motion domain module, an image was considered \textit{globally static} if the dominant value of optical flow magnitude was smaller than $t_{f}=15px$. A minimal value of PSNR calculated within well-exposed mask for the \textit{static} patch to be considered as supervision was set to $45dB$. Values of thresholds for considering a patch as well-exposed were set the same as in the exposure domain module. A \textit{pseudo-static} movement augmentation was set to be drawn from Gaussian distribution: $\mathcal{N}(0, 4)$, while for larger movements: $\pm\mathcal{N}(20, 3)$. Thereafter, we obtained 4 subsets from each processed dataset:
\begin{enumerate}[itemsep=1pt,topsep=2pt,parsep=0pt]
    \item exposure domain \textit{pseudo-static} 
    - ED,
    \item exposure domain with bigger movement 
    - EDM,
    \item motion domain \textit{pseudo-static} 
    - MD,
    \item motion domain  with bigger movement 
    - MDM.
\end{enumerate}

We show in Table \ref{tab:stats} a breakdown of the number of supervision pairs generated for each dataset, considering each domain split.
\begin{table}
\vspace{-2em}
\begin{center}
\caption{A summary of the number of supervision pairs generated for different domain splits for various datasets. Presented as a percentage of the total number (in \textit{italic}) of pairs in the given dataset. In \textbf{bold} - subsets used for testing.}
\vspace{1em}
\label{tab:stats}
\begin{tabularx}{0.95\textwidth}{l|YY|Y|YYYYY|Y}
\toprule
Subset & KTr & \textcolor{black}{\textbf{KTe}} & Sum & \textcolor{black}{\textbf{D}}  & DnGT & S & \textcolor{black}{\textbf{SRM}} & HdM2 & Sum\\
\midrule
ED & 24.11 & 4.47 & 28.58 & 5.49 & 7.27 & 5.14 & 4.57 & 0.54 & 23.02 \\
EDM & 23.18 & 4.16 & 27.34 & 5.61 & 7.15 & 5.00 & 4.39 & 0.52 & 22.67 \\
MD & 18.76 & 3.74 & 22.50 & 3.61 & 9.57 & 6.51 & 6.23 & 0.27 & 26.19 \\
MDM & 17.88 & 3.70 & 21.57 & 3.50 & 11.81 & 6.38 & 6.18 & 0.27 & 28.12 \\
\midrule
Total & 83.93 & 16.07 & \textit{39382} & 18.21 & 35.80 & 23.03 & 21.37 & 1.59 & \textit{106916} \\
\bottomrule
\end{tabularx}
\end{center}
\vspace{-3em}
\end{table}

\subsection{Results}

We report in Table \ref{tab:results} and \ref{tab:results_chen} the results of our approach compared against other HDR estimation approaches, including weakly-, and unsupervised methods for Kalantari~\cite{Kalantari17} and Chen \etal\cite{Chen21} datasets respectively.

\begin{table}
\vspace{-2em}
\begin{center}
\caption{Quantitative comparison of our method against existing approaches on Kalantari \etal dataset. Table is split based on the level of supervision in the respective method: left (\textbf{S}) - fully-supervised, top right (\textbf{WS}) - weakly-supervised, bottom right (\textbf{US}) - unsupervised. The best unsupervised score is highlighted in \textbf{bold}, the best score overall - \underline{underlined}. \textdagger Values as reported in \cite{Wang21}.}
\vspace{1em}
\label{tab:results}
\begin{tabular}{@{}llcccc@{}llccc}
\cmidrule[\heavyrulewidth](){1-5}\cmidrule[\heavyrulewidth](){7-11}
 & & \multicolumn{3}{c}{Kalantari - KTe} & \ \, &  & & \multicolumn{3}{c}{Kalantari - KTe}\\
 \cmidrule(lr){3-5}\cmidrule(lr){9-11}
&  & $P_L$ & $P_{\mu}$ & $HV2$ & & & & $P_L$ & $P_{\mu}$ & $HV2$\\
\cmidrule(){1-5}\cmidrule(){7-11}    
\multirow{7}{*}{\textbf{S}} & AHDR\cite{Yan19} & 41.16 & \underline{43.57} & 64.83 & & \multirow{2}{*}{\textbf{WS}} & Prab-5s\cite{Prabhakar21} & 41.28 & 41.67  &  65.15 \\
& Kalantari\cite{Kalantari17}  & 41.23 & 42.70  & 64.63 & & & Prab-1s\cite{Prabhakar21} & 41.03 & 41.22  &  64.61 \\
\specialrule{0pt}{0pt}{-.4ex}
\cmidrule(){7-11}
\specialrule{0pt}{-.65ex}{0pt}
& Wu\cite{Wu18}  & 41.62 & 42.01   & \underline{65.78} & & \multirow{5}{*}{\textbf{US}} & Prab-0s\cite{Prabhakar21} & \textbf{40.90} & 41.14  & \textbf{64.89} \\
& Prabhakar \cite{Prabhakar19}  & 40.31 & 42.79  & 62.95 & & & Hu\textsuperscript{\textdagger}\cite{Hu13} & 30.84 & 32.19 & 55.25\\
& Prab-SV\cite{Prabhakar21} & \underline{41.79} & 41.94 & 65.30 & & & Oh\textsuperscript{\textdagger}\cite{Oh15} & 27.11 & 27.35 & 46.83\\ 
& & & & & & & Sen \cite{Sen12} & 38.38  & 40.98 & 60.54 \\
& Ours (Supervised) & 40.83 & 42.39  & 64.20 &  & & Ours & 40.54 & \textbf{42.15}  & 63.99\\
\cmidrule[\heavyrulewidth](){1-5}\cmidrule[\heavyrulewidth](){7-11}
\end{tabular}
\end{center}
\vspace{-2em}
\end{table}

We test the proposed method on Kalantari \etal  \cite{Kalantari17} Test split (KTe), while training on self-supervision pairs from the Training and Test splits (KTr+KTe). Additionally, we provide results on Chen dynamic dataset (D) and static dataset with random synthetic movement (SRM), trained on full data introduced by the authors (D+DnGT+S+SRM+HdM2). We provide values of PSNR, PSNR for $\mu$-law tonemapped images, and HDR-VDP2 \cite{Mantiuk11} averaged across all the images in the dataset. All our results were obtained by training the aforementioned UNet-based model for $300$ epochs, setting $\alpha=0.2$, using Adam optimiser \cite{Diederik15} with learning rate $1e-4$, and multi-step scheduler decreasing learning rate by $90\%$ in epochs $210$ and $285$. 

We compare our method against recent self-supervised and unsupervised approaches, \ie~Sen \cite{Sen12}, Hu \cite{Hu13}, Oh \cite{Oh15}, Prabhakar \cite{Prabhakar21} in their proposed zero-shot setting. Additionally, we show the results of weakly-supervised method - Prabhakar \cite{Prabhakar21} with the supervision from 5 static, and 1 or 5 dynamic scenes with ground truth labels. Additionally, we compare ourselves to Prabhakar \cite{Prabhakar21} supervised, Chen \cite{Chen21}, AHDR \cite{Yan19}, Kalantari \cite{Kalantari17}, Wu \cite{Wu18}, Prabhakar \cite{Prabhakar19}.

\begin{table}
\vspace{-1em}
\setlength{\tabcolsep}{4pt}
\begin{center}
\caption{Quantitative comparison of our method against existing approaches on Chen \etal datasets. Table is split based on the level of supervision in the respective method: top - fully-supervised, middle - weakly-supervised, bottom - unsupervised. The best unsupervised score is highlighted in \textbf{bold}, the best score overall - \underline{underlined}.}
\vspace{1em}
\label{tab:results_chen}
\begin{tabular}{@{}lcccccc}
\toprule
 & \multicolumn{3}{c}{Chen - D} & \multicolumn{3}{c}{Chen - SRM}\\
 \cmidrule(lr){2-4}\cmidrule(lr){5-7}
 & $P_L$ & $P_{\mu}$ & $HV2$ & $P_L$ & $P_{\mu}$ & $HV2$ \\
\midrule        
 AHDR\cite{Yan19} & 35.09 & 39.56  & 63.78 & 35.49 & 35.48 & 61.31 \\
Chen\cite{Chen21} & 42.33 & 41.47  & \underline{71.87}  & 41.02 & 35.89 & 68.25\\
Prab-SV\cite{Prabhakar21} & 38.19 & 40.89 & 65.48 & 40.27 & 37.61  & 67.46 \\
\hline
Prab-5s\cite{Prabhakar21}  & 38.90 & 40.02  & 67.51 & 40.42  & 35.86  &  67.31 \\
Prab-1s\cite{Prabhakar21} & 39.29 & 39.65  & 67.05 & 39.74  & 35.08  &  66.93\\
\hline
Prab-0s\cite{Prabhakar21} & 39.01 & 39.51  & 69.12 & 40.52  & 35.04  & 69.97 \\
Sen\cite{Sen12}  & 39.58 & 40.79 & 68.83 & 40.51 & 36.83 &  66.01  \\
Ours & \underline{\textbf{42.80}} & \underline{\textbf{42.05}} & \textbf{71.55} & \underline{\textbf{45.90}} & \underline{\textbf{40.71}} & \underline{\textbf{71.72}}\\
\bottomrule
\end{tabular}
\end{center}
\vspace{-2em}
\end{table}

We test the proposed method on Kalantari \etal  \cite{Kalantari17} Test split (KTe), while training on self-supervision pairs from the Training and Test splits (KTr+KTe). Additionally, we provide results on Chen dynamic dataset (D) and static dataset with random synthetic movement (SRM), trained on full data introduced by the authors (D+DnGT+S+SRM+HdM2). We provide values of PSNR, PSNR for $\mu$-law tonemapped images, and HDR-VDP2 \cite{Mantiuk11} averaged across all the images in the dataset. All our results were obtained by training the 
UNet-based model for $300$ epochs, with $\alpha=0.2$, using Adam optimiser \cite{Diederik15} with learning rate $1e-4$, and multi-step scheduler decreasing learning rate by $90\%$ in epochs $210$ and $285$. 

We compare our method against recent self-supervised and unsupervised approaches, \ie~Sen \cite{Sen12}, Hu \cite{Hu13}, Oh \cite{Oh15}, Prabhakar \cite{Prabhakar21} in their proposed zero-shot setting. Additionally, we show the results of weakly-supervised method - Prabhakar \cite{Prabhakar21} with the supervision from 5 static, and 1 or 5 dynamic scenes with ground truth labels. Additionally, we compare ourselves to Prabhakar \cite{Prabhakar21} supervised, Chen \cite{Chen21}, AHDR \cite{Yan19}, Kalantari \cite{Kalantari17}, Wu \cite{Wu18}, Prabhakar \cite{Prabhakar19}.

Our method outperforms other methods that do not require supervision. Note that zero-shot experiment of Prabhakar \etal~\cite{Prabhakar21} still requires providing 5 completely static scenes. Our approach does not require any assumption on any number of data. Additionally, we achieve a significant improvement over the non-learning approach by Sen \etal \cite{Sen12}. Further, our algorithm provides results comparable to those of fully supervised methods. For Chen \etal dataset, we do outperform other fully-supervised methods trained on the mentioned dataset ($0.47dB$ improvement on dynamic dataset, and $4.88dB$ on static with random motion over the runner-up).
We attribute such a good performance to the ability of our self-supervised method to facilitate in-domain data (D and SRM) which are not presented to the network in supervised methods.
We show that our approach provides an HDR estimation of quality superior to other unsupervised methods, and competitive with supervised ones, even though we do not use any annotated data and use a simple, lightweight vanilla architecture. Additionally, we would like to highlight that our proposed self-supervised strategy obtains very close performance to the same architecture trained in a fully supervised setting (Table \ref{tab:results_chen} Ours (Supervised) vs Ours), which further validates the effectiveness of our HDR pseudolabels.

\begin{figure*}
    \centering
    \includegraphics[width=0.98\linewidth]{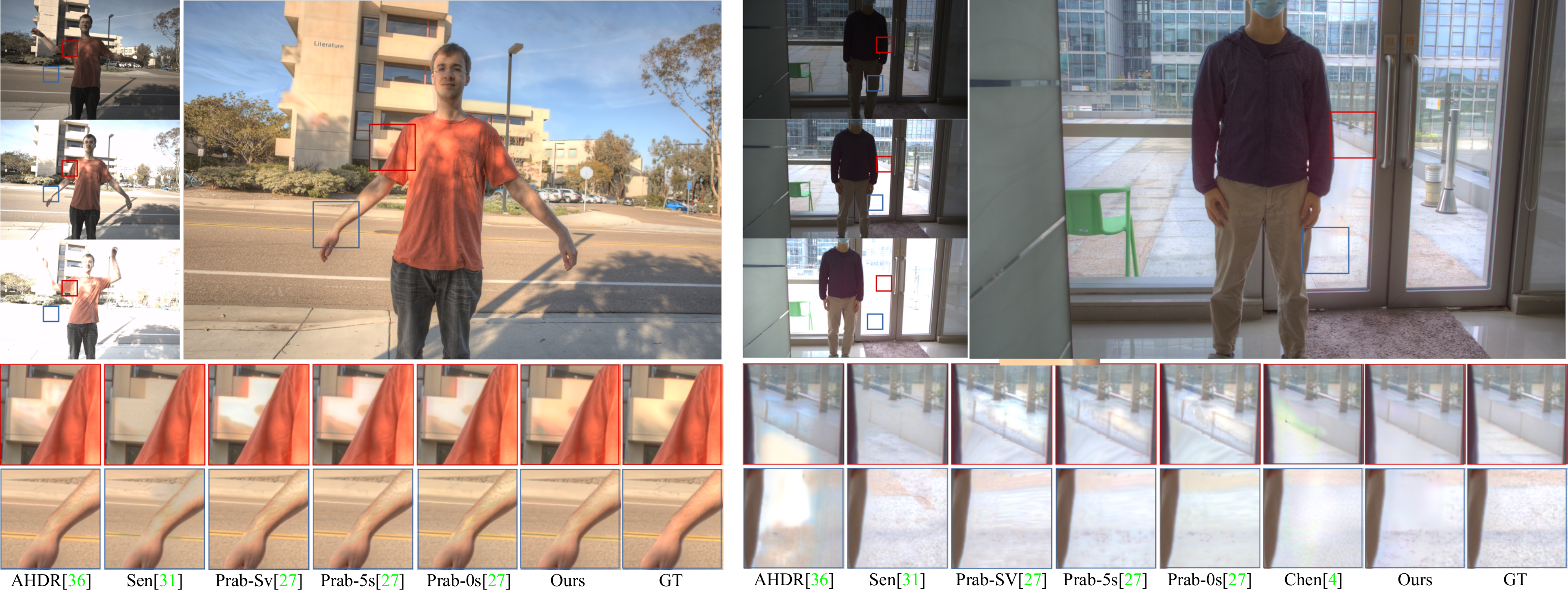}
    \caption{Qualitative comparison of our method against fully-, weakly-, and unsupervised approaches. Left image - example from Kalantari \etal , note artefacts in colour reconstruction in red, hand discolouration in blue. Right image - example from Chen \etal Dynamic, note artefacts on the wall in red, pavement ghosting in blue.}
    \label{fig:results}
\vspace{-1em}
\end{figure*}

Figure \ref{fig:results} presents a qualitative results of our method on an image from Kalantari \etal and an image from Chen \etal - Dynamic compared against main methods reported in Table \ref{tab:results}. In the red patch of the left image, note the discolouration of the building observable in various methods, while reconstructed well in ours - we attribute this to supervision from images with synthetically increased illumination. Similarly, blue-marked patch present skin-tone colour reconstruction other methods are struggling with, leaking the yellow colour from road marks. In the right image, red patch focuses on various artefacts on the highly illuminated in the reference image, wall. Even though some slight ghosting is visible, our method provides a good reconstruction of that region, keeping its colour correct, and uniform. Blue patch accounts for ghosting effects present in most of the methods in that region. We argue that our method and Sen \cite{Sen12} provide the most visually satisfying reconstruction of the pavement plane, with the latter, however, introducing new lines not present in the ground truth.

\subsection{Ablation study}

\begin{figure*}
\vspace{-1em}
    \centering
    \includegraphics[width=0.84\linewidth]{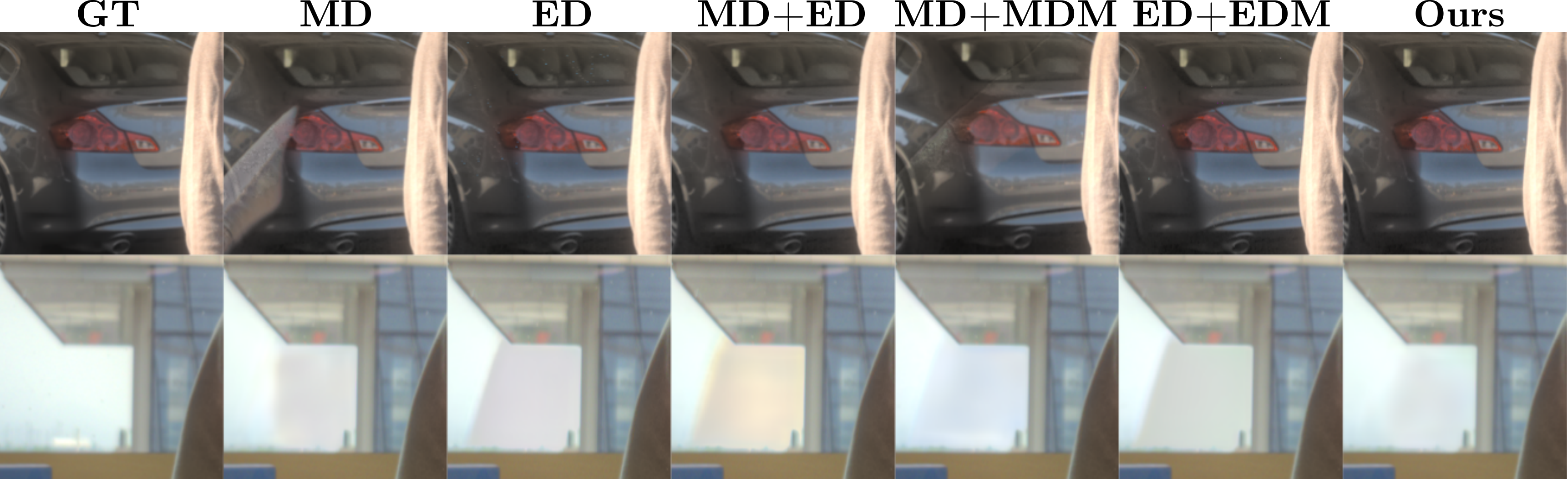}
    \caption{
    Qualitative results for domain split ablation. \textbf{Top:} Note the ghosting artifacts from high illumination long photo, when trained only with motion domain. \textbf{Bottom:} Observe inaccuracies of HDR fusion when trained only in exposure domain.}
    \label{fig:ablation}
\vspace{-1em}
\end{figure*}

In Table \ref{tab:ablation_domains} we emphasise the importance of each module in creating a self-supervision signal for network training. We perform an ablation study by removing a single block from the pipeline in each experiment, and training only on non-augmented versions of motion and exposure domains. 
\begin{table}
\vspace{-1em}
\small
\begin{center}
\caption{Ablation results obtained by training with different domain splits of datasets for Kalantari \etal and Chen \etal.}
\vspace{1em}
\label{tab:ablation_domains}
\begin{tabularx}{0.65\textwidth}{@{}l@{}@{}rYYYYYY}
\toprule
 & MD  & \cmark & \xmark & \cmark & \cmark & \xmark & \cmark \\
 & MDM & \xmark & \xmark & \xmark & \cmark & \xmark & \cmark \\
 & ED  & \xmark & \cmark & \cmark & \xmark & \cmark & \cmark \\
 & EDM & \xmark & \xmark & \xmark & \xmark & \cmark & \cmark \\
\midrule
\multirow{2}{*}{KTe} & $P_L$ & 36.44 & 39.55 & 38.80 & 37.65 & 40.40 & 40.54\\
& $P_{\mu}$ & 36.73 & 41.16 & 38.50 & 37.76 & 41.68 & 42.15\\
\midrule
\multirow{2}{*}{D} & $P_L$ & 38.72 & 38.38 & 40.09 & 41.34 & 40.22 & 42.80\\
& $P_{\mu}$ & 39.40 & 39.15 & 39.25 & 41.62 & 39.70 & 42.05 \\
\midrule
\multirow{2}{*}{SRM} & $P_L$ & 45.44 & 42.50 & 46.01 & 43.99  & 41.93 & 45.90\\
& $P_{\mu}$  & 41.45 & 35.04 & 40.10 & 41.48 & 34.98  & 40.71\\
\bottomrule
\end{tabularx}
\end{center}
\vspace{-2em}
\end{table}
It is worth to notice that training in augmented motion domain (MD+MDM) only achieved a very good performance on both Chen \etal datasets, whereas training in exposure domain (ED+EDM) performed significantly worse. On the other hand, exposure domain supervision performed almost as good as the whole supervision set for Kalantari \etal dataset (both ED, and ED+EDM), while motion domain supervision performance was noticeably degraded. This, in conjunction with full supervision signal achieving the highest results proves the importance of exploiting both domains for self-supervision. Additionally, motion augmentation always improves the results over single ED or MD on dynamic datasets showing that non-augmented splits may not have enough motion samples. Qualitative comparison is also present in Figure \ref{fig:ablation}. We can observe the network struggling with ghosting artefacts when not trained with any exposure domain split. With lack of motion domain split, we notice a decreased capability in fusing the HDR images. We observe a non-motion-augmented training to be the most effective for non-dynamic dataset (Chen \etal - static with random movement). However, presenting the network with higher values of misalignment in the images is shown to be important for improving the results on real, dynamic data.

Table \ref{tab:ablation_datasets} focuses on understanding the behaviour of our strategy when using various datasets for training. We can observe a good performance on both Chen \etal datasets when training the network only with patches extracted from the respective set of images, or from both Chen \etal test datasets combined. The results for Kalantari show, however, a significant drop in performance when trained only on test set patches. We hypothesise that such a worsening of the results might be caused by the reduced size of the dataset, and thus the reduced amount of supervision patches extracted from Kalantari \etal Test (KTe, see Table \ref{tab:stats}). This suggests a high capability of our method when used only within the given set of data, provided a sufficient number of input images. Additionally, the results of training our method with non-target dataset, and in particular full sets of data prove its usefulness in incorporating additional, not expensive to obtain, unlabelled series of images into the training pipeline. Similarly, performance of the models trained on Chen \etal datasets is reported to be higher on Kalantari \etal Test than the model trained on the respective dataset. We attribute this behaviour to the larger number of patches extracted from Chen \etal datasets. Additionally, this proves usefulness of our method by presenting the transferability between various image domains.

Finally, in Table \ref{tab:ablation_arch} we present a comparison of results of training with various architectures. We substitute the backbone of our approach (UNet \cite{Ronneberger15}) with (i) a
ResNet backbone, as described in Zhang \etal \cite{Zhang18} (with
no dense connections); and (ii) a Grouped Residual Dense
Blocks architecture as described in Kim \etal \cite{Kim19} (GRDN). We show that the method we propose is suitable to be used with various neural networks. Additionally, we note that increasing the capacity of the network (GDRN characterises with high complexity) yields increase in performance when provided with enough data (number of patches generated from Chen \etal is significantly greater than number of patches obtained from Kalantari \etal dataset).

\begin{table}
\vspace{-2em}
\small
\begin{minipage}{0.56\linewidth}
    \begin{center}
    \caption{Ablation results obtained by networks trained on different datasets. In \textbf{bold} patches only from dataset used for testing.}
    \label{tab:ablation_datasets}
    \end{center}
\end{minipage}
\begin{minipage}{0.42\linewidth}
    \begin{center}
    \caption{Comparison of quantitative results on Chen\cite{Chen21} D and SRM for various backbone architectures.}
    \label{tab:ablation_arch}
    \end{center}
\end{minipage}
\begin{minipage}{\linewidth}
    \vspace{1em}
    \begin{tabularx}{\textwidth}{rccYYYYcrcYYY}
    \cmidrule[\heavyrulewidth](){1-7}\cmidrule[\heavyrulewidth](){9-13}
    \multicolumn{2}{c}{{\scriptsize Test\textbackslash Train}} & Ours & KTe & D & SRM & D\texttt{+}SRM & \ \, &  &  & UNet & ResNet & GRDN\\
    \cmidrule(){1-7}\cmidrule(){9-13}
    \multirow{2}{*}{KTe} & $P_L$ & 40.54 & \textbf{37.09} & 38.86 & 38.48 & 38.72 & &  \multirow{3}{*}{D} & $P_L$ & 42.80 & 41.64 & 44.36\\
    & $P_{\mu}$ & 42.15 & \textbf{37.76} & 41.49 & 41.31 & 40.30 & & &  $P_{\mu}$ & 42.06 & 42.20 & 43.30\\
    \multirow{2}{*}{D} & $P_L$ & 42.80 & 35.48  & \textbf{41.11} & 39.01 & 41.19 & & & $HV2$ & 71.55 & 70.76 & 73.77\\
    & $P_{\mu}$ & 42.05 & 38.74 & \textbf{40.03} & 41.54 & 40.95 & & \multirow{3}{*}{SRM} & $P_L$ & 45.90 & 43.34 & 45.97\\
    \multirow{2}{*}{SRM} & $P_L$ & 45.90 & 36.58 & 39.71 & \textbf{44.67} & 45.48 & &  & $P_{\mu}$ & 40.71 & 39.93 & 42.39\\
    & $P_{\mu}$ & 40.71 & 34.64 & 35.09 & \textbf{40.65} & 39.48 & & & $HV2$ & 71.72 & 68.88 & 70.46\\
    \cmidrule[\heavyrulewidth](){1-7}\cmidrule[\heavyrulewidth](){9-13}
    \end{tabularx}
\end{minipage}

\vspace{-3em}
\end{table}

\section{Conclusions}

In this work we propose a self-supervised approach for HDR Imaging that exploits LDR patches based on its motion and exposure characteristics. Our method provides supervision pairs for data-driven algorithms obtained only from sets of LDR images captured without any constrains. We have shown that the supervision provided by our approach enables training networks with performance superior to other weakly- and unsupervised approaches, and comparable to supervised ones. Our strategy can leverage any image sequence captured with varying exposure to increase the accuracy of HDR estimation, saving a lot of time and effort on capturing training data. Additionally we provide an effective way to incorporate unlabelled in-domain data into the training process. 

Our work provides a simple yet effective model for applying synthetic illumination gain to the image. A further step to develop supervision in the exposure domain would be to study the possibility of learning that degradation model, \eg via training a GAN network with a set of unpaired saturated patches. 

\clearpage
\appendix
\vspace*{\stretch{0.5}}
\begin{center}
    \Large\textbf{Supplementary material}
\end{center}
\vspace*{\stretch{1.0}}

\section{Generalisation Experiments}
In order to show generalisability of our approach, we performed a comparison of various network architectures used for the HDR reconstruction training and inference (Table \ref{tab:ablation_arch} and \ref{tab:results_sup_kal}). In Table \ref{tab:params} we present the comparison of the number of parameters and number of floating point operations per second (GFLOPs) for a single test image of size $1000\mathrm{x}1500$~px.
The ResNet architecture presents a lower number of parameters, however slightly higher computational complexity than UNet. GRDN is a much more complex architecture than UNet in terms of both number of parameters and complexity. 

In Table \ref{tab:results_sup_kal} we report the results of various backbones experiments for Kalantari \etal~\cite{Kalantari17} Test as well as both Chen \etal~\cite{Chen21} splits.
In the experiment we can observe a similar performance for all backbone architectures. 
We believe that these results show the generalisation capability of our framework with respect to various backbone architectures with different number of parameters and complexity. We suggest that an increase of the gap in training performance on Chen \etal~\cite{Chen21} for GRDN indicates the sensitivity of complex architectures to the cardinality of a dataset (Chen \etal yielded significantly more patches than Kalantari \etal dataset).

\begin{table}
\vspace{-2em}
    \begin{center}
    \caption{Comparison of quantitative results on Kalantari \cite{Kalantari17} and Chen \cite{Chen21} for various backbone architectures.}
    \label{tab:results_sup_kal}
    \begin{tabularx}{0.9\textwidth}{@{}lYYYYYYYYY}
    \toprule
     & \multicolumn{3}{c}{Kalantari - Te} & \multicolumn{3}{c}{Chen - D} & \multicolumn{3}{c}{Chen - SRM}\\
     \cmidrule(lr){2-4}\cmidrule(lr){5-7}\cmidrule(lr){8-10}
     & $P_L$ & $P_{\mu}$ & $HV2$ & $P_L$ & $P_{\mu}$ & $HV2$ & $P_L$ & $P_{\mu}$ & $HV2$\\
    \midrule        
    Unet & 40.54 & 42.15 & 63.99 & 42.80 & 42.06 & 71.55 & 45.90 & 40.71 & 71.72 \\
    ResNet & 40.45 & 42.18 & 63.40 & 41.64 & 42.20 & 70.76 & 43.34 & 39.93 & 68.88 \\
    GRDN & 40.81 & 42.31 & 64.36 & 44.36 & 43.30 & 73.77 & 45.97 & 42.39 & 70.46 \\
    \bottomrule
    \end{tabularx}
    \end{center}
\vspace{-2em}
\end{table}

\begin{table}
\vspace{-2em}
\begin{center}
\caption{Number of parameters and GFLOPs comparison for various backbone architectures.}
\label{tab:params}
\begin{tabular}{lcc}
\toprule
Architecture & Parameters & GFLOPs \\
\midrule
UNet & 17.3M & 1856 \\
ResNet & 0.8M & 2240 \\
GRDN & 21.9M & 6570\\
\bottomrule
\end{tabular}
\end{center}
\vspace{-1em}
\end{table}

\begin{figure}
    \centering
    \includegraphics[width=0.95\linewidth]{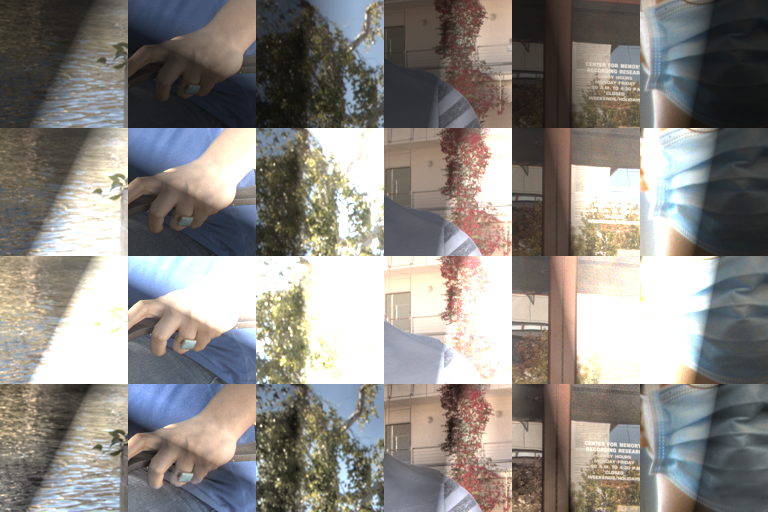}
    \caption{Examples of patches generated with synthetic gain applied progressively from a randomly generated line. Top to bottom: short, medium, long, HDR pseudo-label.}
    \label{fig:patch_example_exp_line}
\end{figure}
\begin{figure}
    \centering
    \includegraphics[width=0.95\linewidth]{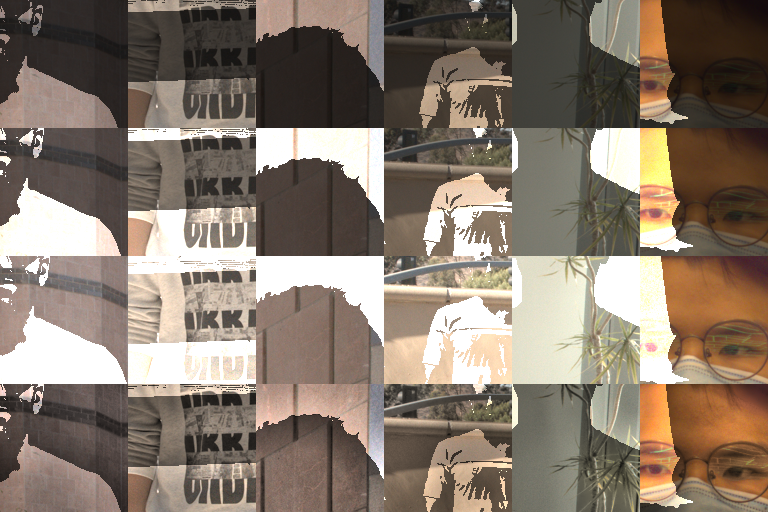}
    \caption{Examples of patches generated with synthetic gain applied by transferring a saturation mask from overexposed patches. Top to bottom: short, medium, long, HDR pseudo-label.}
    \label{fig:patch_example_exp_mask}
\end{figure}
\begin{figure}
    \centering
    \includegraphics[width=0.95\linewidth]{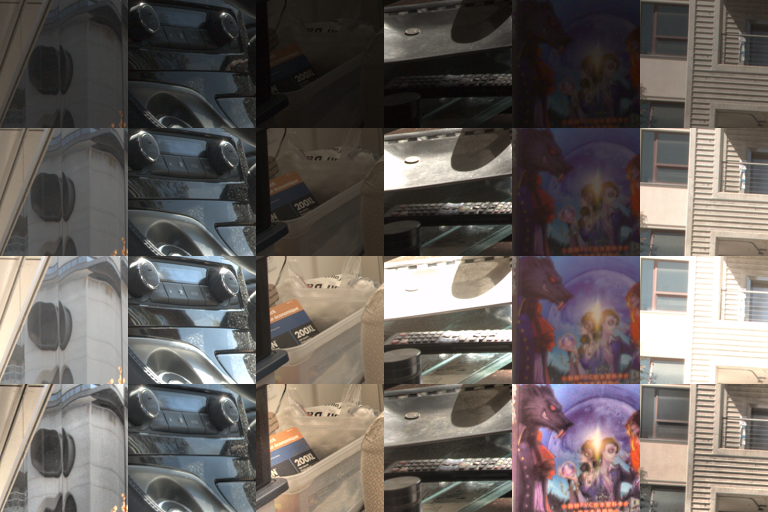}
    \caption{Examples of patches generated as a fusion of static patches. Top to bottom: short, medium, long, HDR pseudo-label.}
    \label{fig:patch_example_stat_fuse}
\end{figure}
\begin{figure}
    \centering
    \includegraphics[width=0.95\linewidth]{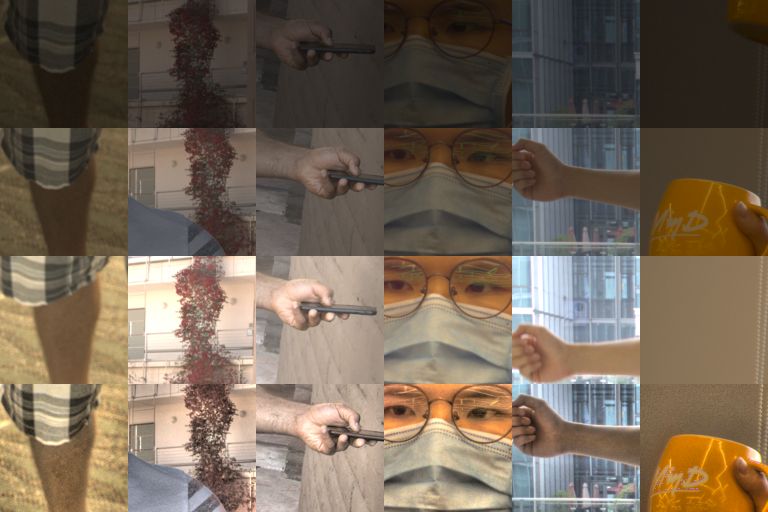}
    \caption{Examples of patches where reference frame is reused as HDR. Top to bottom: short, medium, long, HDR pseudo-label.}
    \label{fig:patch_example_stat_exp}
\end{figure}

\section{Extended Visualisations}

We provide additional examples of patches generated with our approach. Figure \ref{fig:patch_example_exp_line} presents a set of images generated with synthetic gain mask applied progressively from a randomly generated line.
Figure \ref{fig:patch_example_exp_mask} corresponds to supervision pairs in which the synthetic gain mask is transferred from saturated patches. Figure~\ref{fig:patch_example_stat_fuse} shows examples of patches generated via the fusion of static patches.
Figure \ref{fig:patch_example_stat_exp} presents patches with reference frame reused as HDR supervision.
Note that all HDR pseudo-labels were tonemapped for visualisation.

\begin{figure*}[h]
    \centering
    \includegraphics[width=0.9\linewidth]{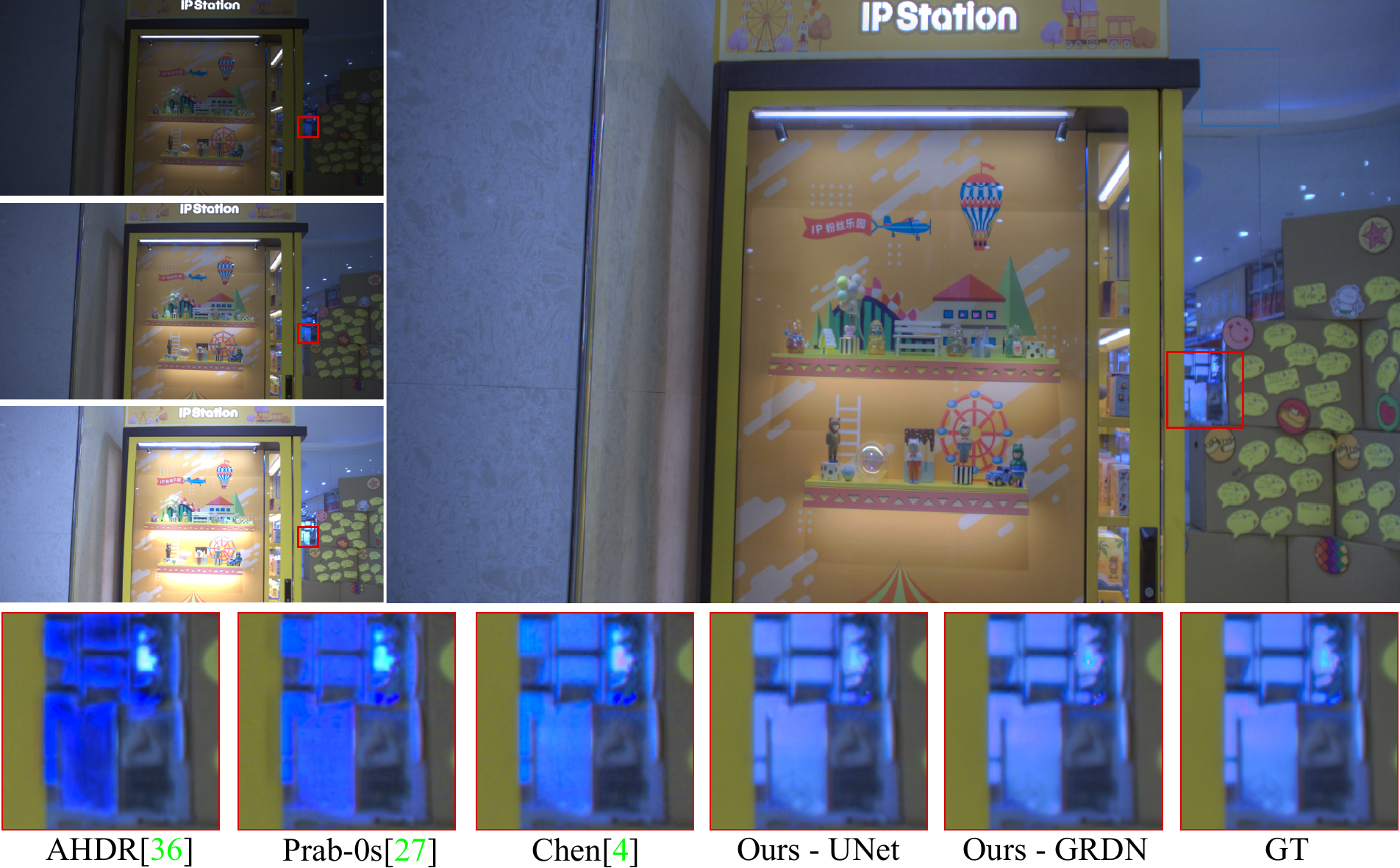}
    \caption{Qualitative comparison of our method against fully-, and unsupervised approaches, including generalisation experiment. Example from Chen \etal.}
    \label{fig:results_sup}
    \vspace{-2em}
\end{figure*}

Figure \ref{fig:results_sup} presents additional qualitative results of our method, including examples obtained in generalisation experiment (GRDN), unsupervised approach from Prabhakar \cite{Prabhakar21} (zero-shot setting), and supervised methods: Chen \etal \cite{Chen21}, and AHDR \cite{Yan19}.

\section{Implementation details} 
\subsection{Motion Domain}

\begin{figure*}
    \centering
    \begin{subfigure}{0.32\linewidth}
        \includegraphics[width=0.99\linewidth]{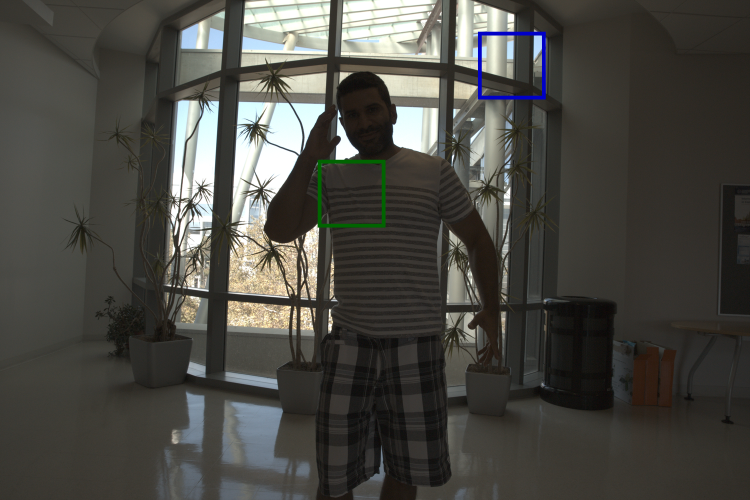}
        \caption{Short.}
    \end{subfigure}
    \begin{subfigure}{0.32\linewidth}
        \includegraphics[width=0.99\linewidth]{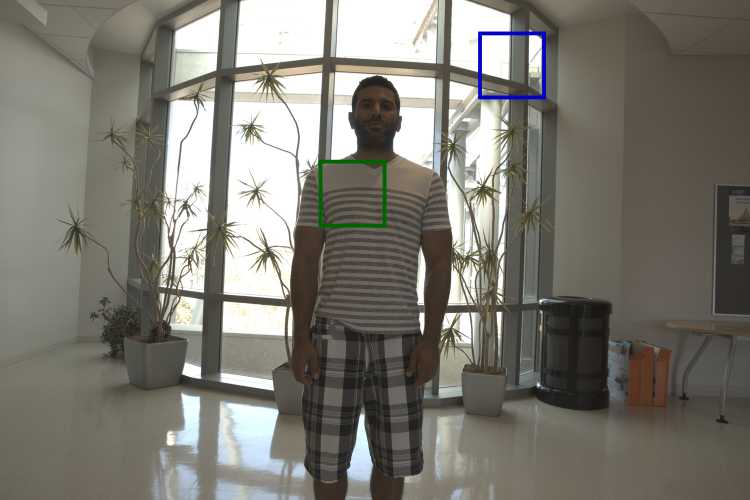}
        \caption{Medium.}
    \end{subfigure}
    \begin{subfigure}{0.32\linewidth}
        \includegraphics[width=0.99\linewidth]{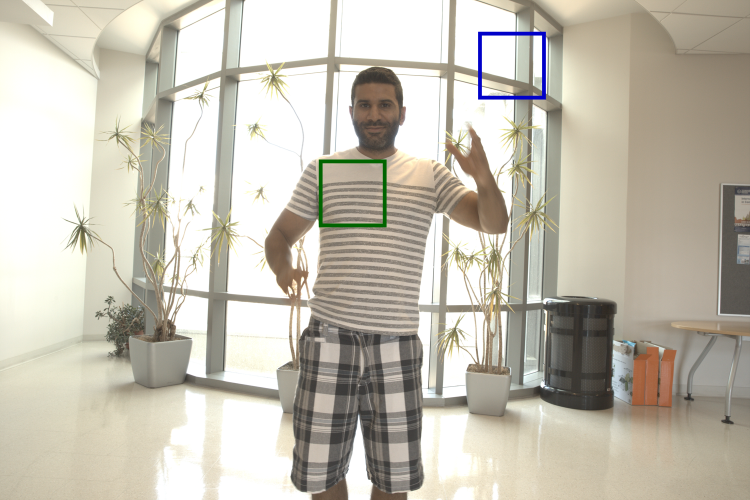}
        \caption{Long.}
    \end{subfigure}
    \vspace{-0.5em}
    \caption{Example of image sequence used for detailed description of patch classification based on motion cues.}
    \label{fig:md_alg_inp}
    \vspace{-0.5em}
\end{figure*}

\begin{figure}
    \centering
    \begin{subfigure}{0.4\linewidth}
        \includegraphics[width=0.99\linewidth]{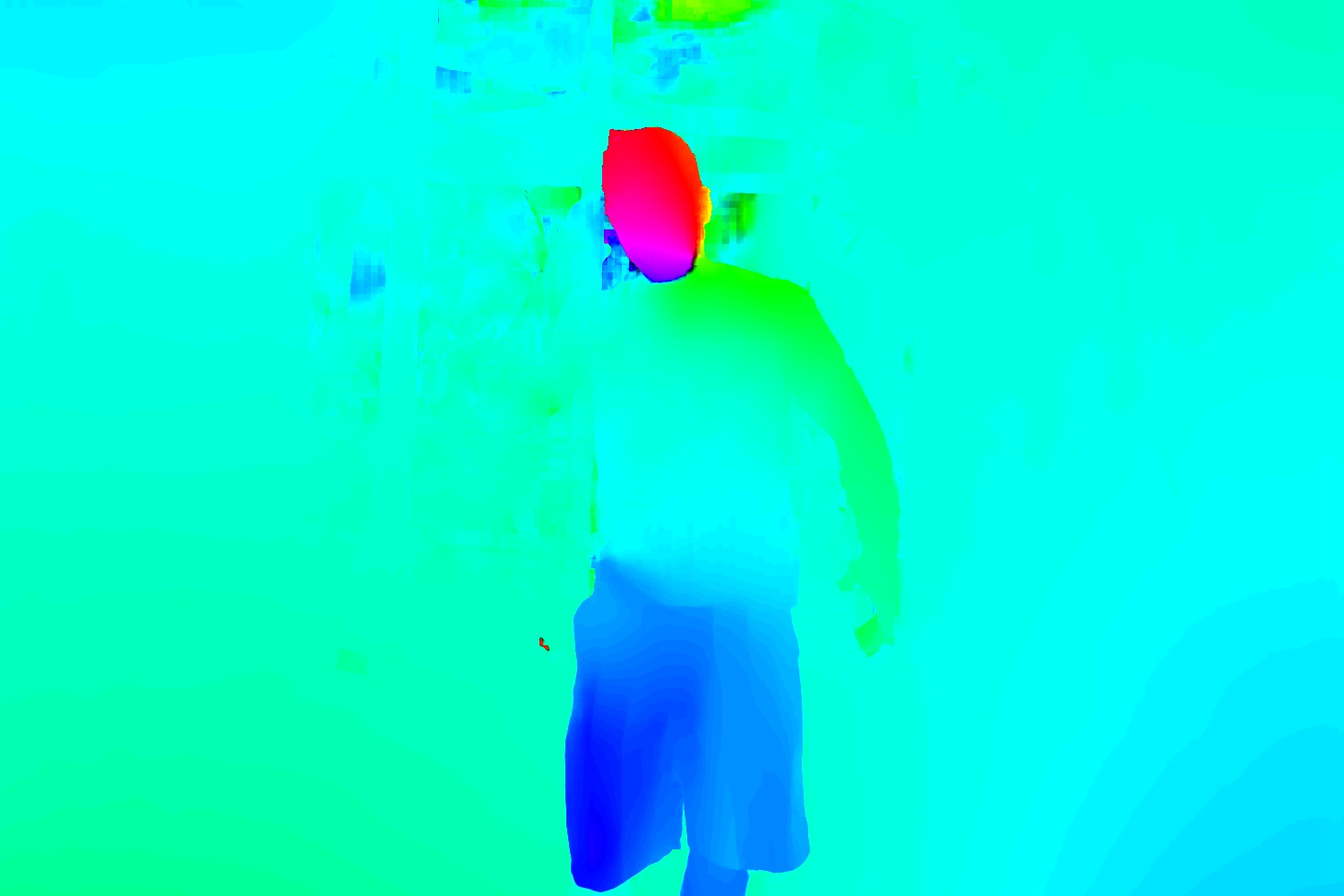}
        \caption{$I_0 \rightarrow I_1$}
    \end{subfigure}
    \begin{subfigure}{0.4\linewidth}
        \includegraphics[width=0.99\linewidth]{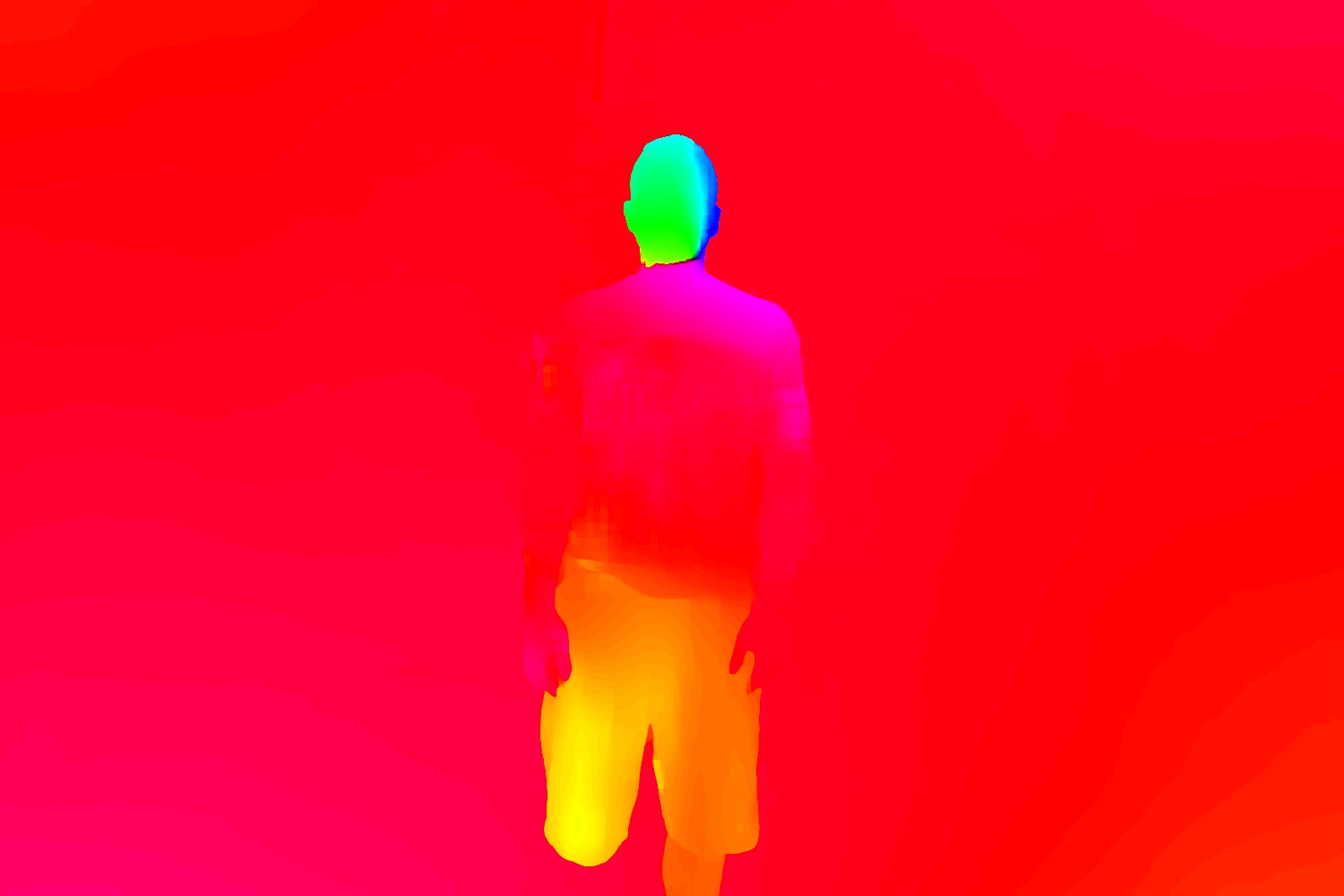}
        \caption{$I_1 \rightarrow I_0$}
    \end{subfigure}
    \begin{subfigure}{0.4\linewidth}
        \includegraphics[width=0.99\linewidth]{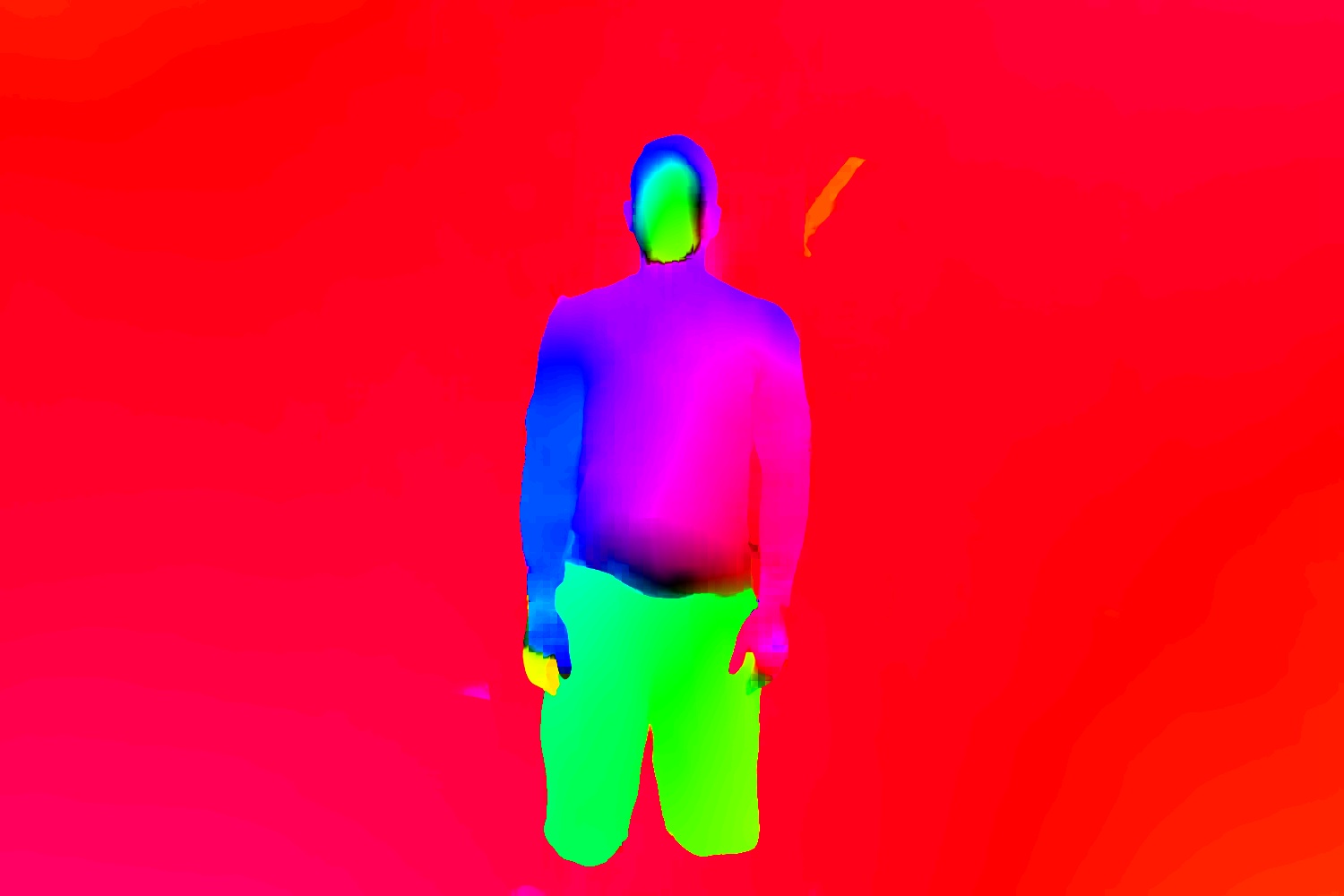}
        \caption{$I_1 \rightarrow I_2$}
    \end{subfigure}
    \begin{subfigure}{0.4\linewidth}
        \includegraphics[width=0.99\linewidth]{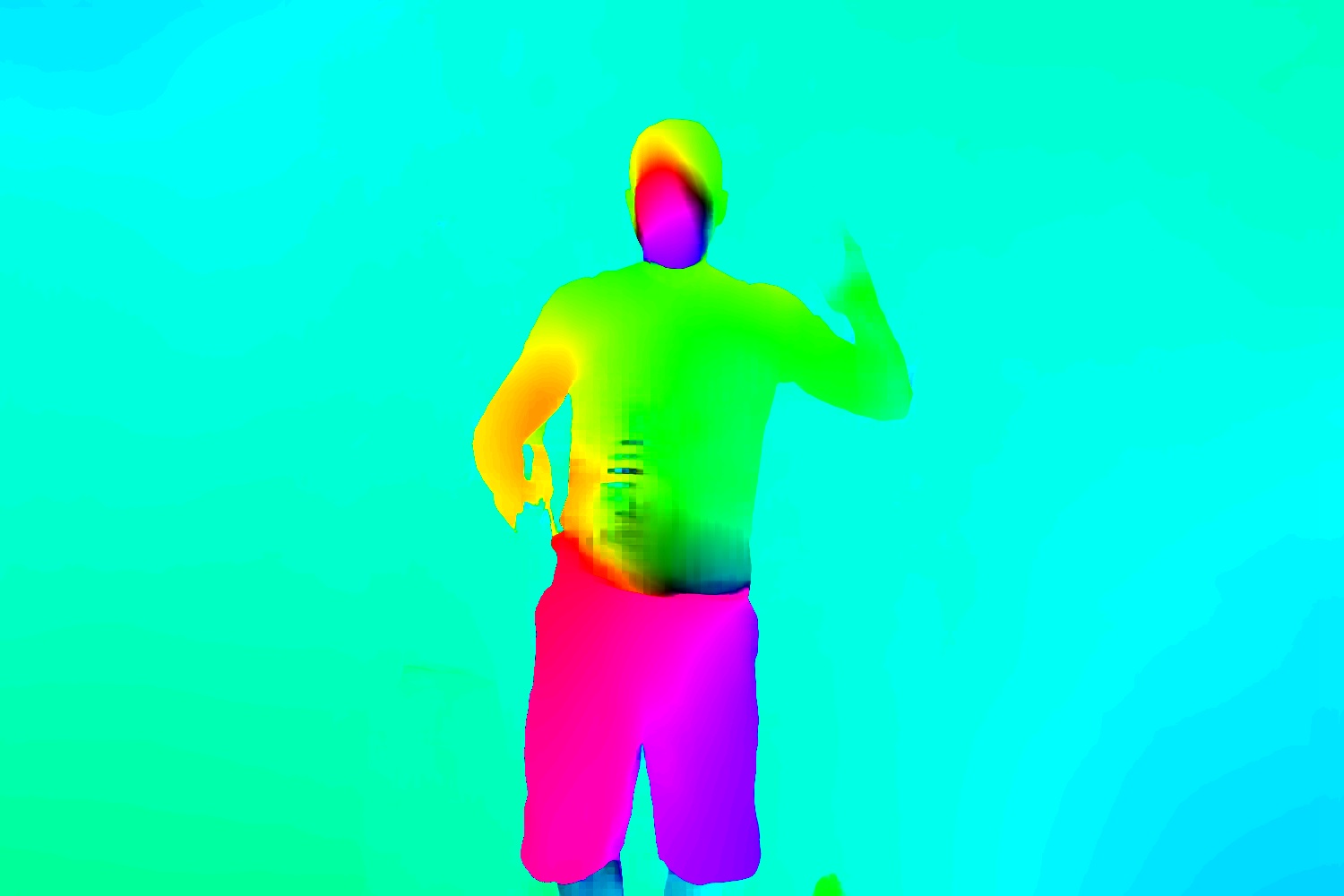}
        \caption{$I_2 \rightarrow I_1$}
    \end{subfigure}
    \vspace{-0.5em}
    \caption{Optical flow map visualisation with $uv$ coordinates mapped to angle as value and magnitude as colour in HSV space.}
    \label{fig:md_alg_of}
    \vspace{-1em}
\end{figure}

In this section we present a detailed explanation of patch classification with respect to motion cues. Firstly, we consider a set of images captured with varying exposure times - $(I_0, I_1, I_2)$, where $I_1$ is the reference frame. An example of such sequence is presented in Figure \ref{fig:md_alg_inp} (note patch colouring, consistent across this section).
We calculate optical flow values for all combinations of images that include a reference frame, \ie $I_1 \rightarrow I_0$, $I_1 \rightarrow I_2$, $I_0 \rightarrow I_1$, $I_2 \rightarrow I_1$. A visualisation of optical flow maps is presented in Figure \ref{fig:md_alg_of}.
\begin{figure}
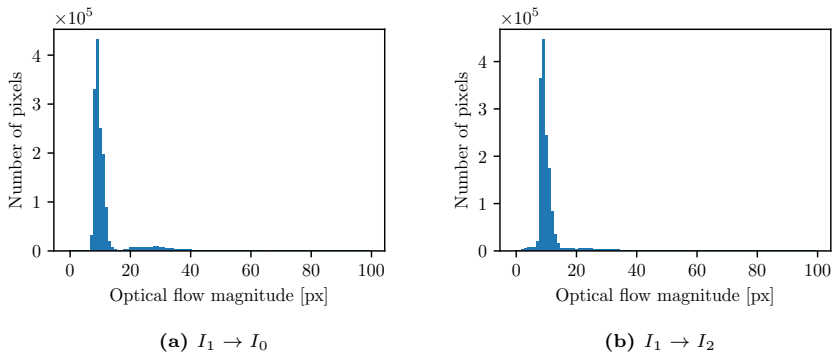

    \centering
    \begin{subfigure}{0.48\linewidth}
        \resizebox{\linewidth}{!}{
            \input{Images_sup/static/flow_1_0_hist.pgf}
        }
        \caption{$I_1 \rightarrow I_0$}
    \end{subfigure}
    \begin{subfigure}{0.48\linewidth}
        \resizebox{\linewidth}{!}{
            \input{Images_sup/static/flow_1_2_hist.pgf}
        }
        \caption{$I_1 \rightarrow I_2$}
    \end{subfigure}
    \caption{Histograms of optical flow magnitude with respect to the reference frame for the considered example.}
    \label{fig:md_alg_histo}
    \vspace{-1em}
\end{figure}
In the next step, we calculate histograms with $1px$ bin width of the magnitude of optical flow for the optical flows calculated with respect to reference frame ($I_1 \rightarrow I_0$, $I_1 \rightarrow I_2$). Such histograms are used to consider whether the images can be aligned based of the threshold discussed in \ref{sec:static_thr}. Histograms obtained for the optical flows related to the given example are presented in Figure~\ref{fig:md_alg_histo}.

\begin{figure*}
    \centering
    \begin{subfigure}{0.32\linewidth}
        \includegraphics[width=0.99\linewidth]{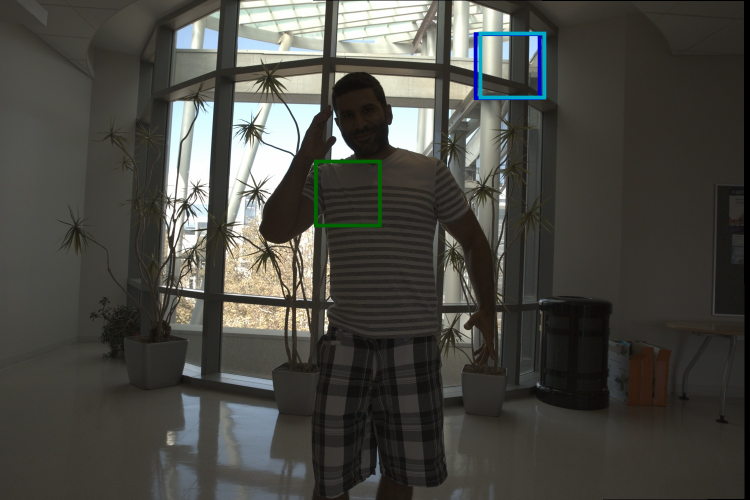}
        \caption{Short.}
    \end{subfigure}
    \begin{subfigure}{0.32\linewidth}
        \includegraphics[width=0.99\linewidth]{Images_sup/static/inp_m.png}
        \caption{Medium.}
    \end{subfigure}
    \begin{subfigure}{0.32\linewidth}
        \includegraphics[width=0.99\linewidth]{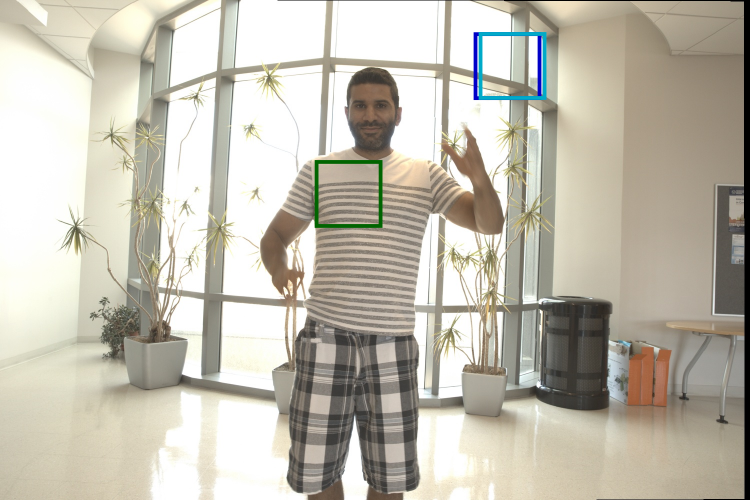}
        \caption{Long.}
    \end{subfigure}
    \caption{Image sequence after applying the warping. Pale blue patch corresponds to the warping of dark blue patch.}
    \label{fig:md_alg_warp}
    \vspace{-1em}
\end{figure*}

If the dominant value in the histogram is higher that the threshold $t_f$, then image is considered to have high misalignment which is not corrected, and its patches are considered as in section 3.1 of main manuscript (\textit{dynamic} patches). For other images, we consider the optical flow estimated displacements to correct the perspective. However, we are not trying to correct dynamic behaviour of the objects in the scene, only camera movement. Therefore, we consider only pixel displacements of dominant and neighbouring bins, and use $\mathtt{RANSAC}$ to estimate homography matrices between $I_0$ and $I_1$ -- $H_{01}$, and $I_2$ and $I_1$ -- $H_{21}$. Further we apply the homographies to non-reference images to obtain aligned images - $I_{0W}$ and $I_{2W}$. Figure \ref{fig:md_alg_warp} presents given example after warping - note that corresponding patches in warped images are aligned, whereas, in original input, alignment is not guaranteed (in fact, extremely rare).
Further, we consider images on the patch level - a set of patches $\{P_{0i}, P_{1i}, P_{2i}, P_{0Wi}, P_{2Wi}\}$ coming from $I_0, I_1, I_2, I_{0W}, I_{2W}$ respectively. For each patch, we test whether its content is static. Given the patch set, we consider the optical flow magnitude in the corresponding region. The magnitude of the optical flow has to not deviate from its median more than a threshold across the whole patch. The threshold is based on the value of the median $m$: $T=max(min(m, 2), 0.5)$ - it allows for a bigger deviation with bigger displacements. The condition for the optical flow magnitude has to be satisfied for the optical flow calculated between all aforementioned pairs of images to consider patch as \textit{static}.

\begin{figure}
    \centering
    \includegraphics[width=0.85\linewidth]{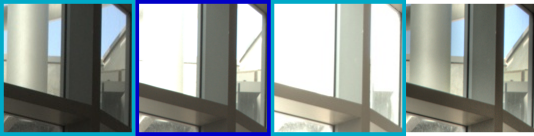}
    \caption{HDR estimation for the \textit{static} patch example. Left to right: short warped, medium, long warped, HDR estimation - tonemapped.}
    \label{fig:md_alg_fusion}
\end{figure}

\begin{figure}
    \centering
    \includegraphics[width=0.85\linewidth]{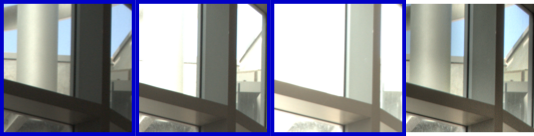}
    \caption{Supervision pair from \textit{static} patch. Left to right: short, medium, long, HDR estimation - tonemapped. Note: input supervision patches are not warped.}
    \label{fig:md_static}
\end{figure}

\begin{figure}
    \centering
    \includegraphics[width=0.85\linewidth]{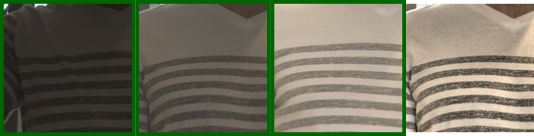}
    \caption{Supervision pair from \textit{dynamic} patch. Left to right: short, medium, long, HDR estimation - tonemapped.}
    \label{fig:md_dynamic}
\end{figure}

\textit{Static} patches are processed as described in Section 3.1 of the main manuscript. Figure \ref{fig:md_alg_fusion} presents a set of warped input patches with resulting HDR pseudo-label for the example given in Figures \ref{fig:md_alg_inp} and \ref{fig:md_alg_warp}. 
Similarly, supervision pair corresponding to the same set of patches is shown in Figure \ref{fig:md_static}.
A supervision pair for a \textit{dynamic} patch for the given example is presented in Figure \ref{fig:md_dynamic}
\begin{figure*}
    \centering
    \includegraphics[width=0.95\linewidth]{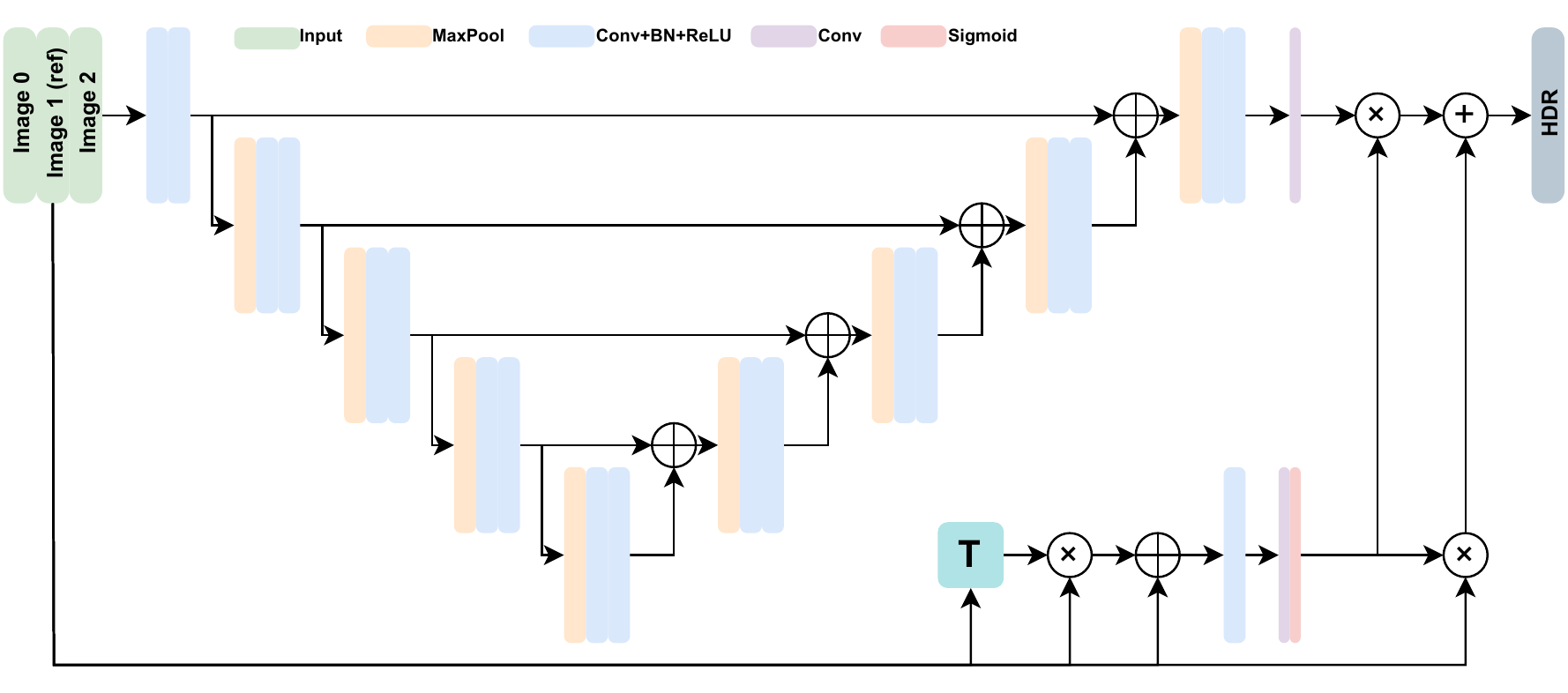}
    \caption{Detailed overview of used UNet architecture.}
    \label{fig:unet_arch}
\end{figure*}

\subsection{UNet Architecture}

In our experiments we used a UNet \cite{Ronneberger15} based model. A detailed overview is presented in Figure \ref{fig:unet_arch}.
We consider a triplet of LDR images as 9 channel input to a regular UNet network that predicts a residual signal to incorporate higher dynamic range in the reference frame. Additionally, we extract a well-exposed mask from a reference image, and input a concatenation of the mask, and masked reference frame (4 channels) to a shallow convolutional branch that predicts blending weights. Finally, HDR estimation is obtained by fusing reference frame with residual signal according to predicted weights.

\subsection{Source Code}
We intend to release the source code together with trained network weights to reproduce the presented results subject to an ongoing internal review and approval.

\section{Hyper-parameters}
\subsection{Static camera threshold} \label{sec:static_thr}
\begin{figure}
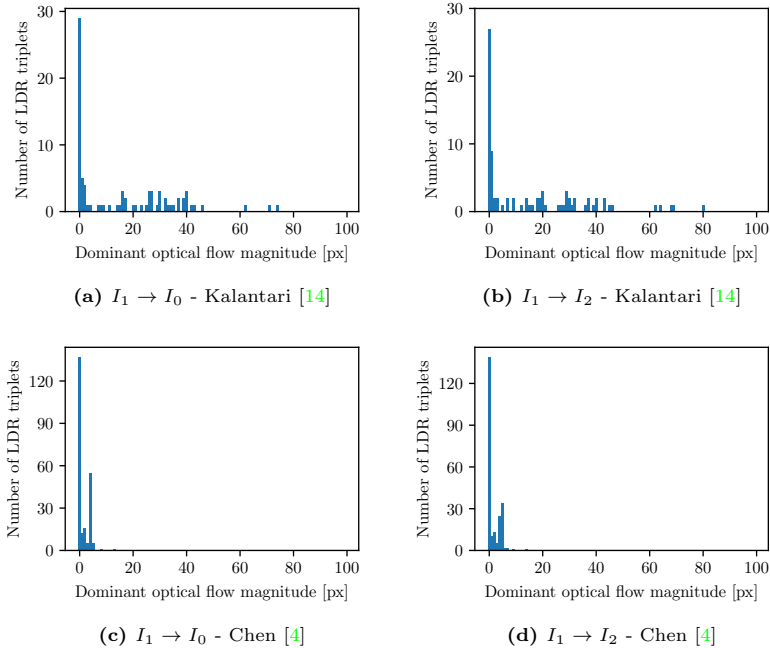

    \vspace{-1em}
    \centering
    \begin{subfigure}{0.44\linewidth}
        \resizebox{\linewidth}{!}{
            \input{Images_sup/kalantari_hist_1_0.pgf}
        }
        \caption{$I_1 \rightarrow I_0$ - Kalantari \cite{Kalantari17}}
    \end{subfigure}
    \begin{subfigure}{0.44\linewidth}
        \resizebox{\linewidth}{!}{
            \input{Images_sup/kalantari_hist_1_2.pgf}
        }
        \caption{$I_1 \rightarrow I_2$ - Kalantari \cite{Kalantari17}}
    \end{subfigure}
    \begin{subfigure}{0.44\linewidth}
        \resizebox{\linewidth}{!}{
            \input{Images_sup/video_hist_1_0.pgf}
        }
        \caption{$I_1 \rightarrow I_0$ - Chen \cite{Chen21}}
    \end{subfigure}
    \begin{subfigure}{0.44\linewidth}
        \resizebox{\linewidth}{!}{
            \input{Images_sup/video_hist_1_2.pgf}
        }
        \caption{$I_1 \rightarrow I_2$ - Chen \cite{Chen21}}
    \end{subfigure}
    \caption{Histograms of dominant optical flow values in Kalantari \cite{Kalantari17} dataset and Chen \cite{Chen21} - dynamic, and static with random movement datasets.}
    \label{fig:threshold_hist}
    \vspace{-1em}
\end{figure}

In our experiments we have chosen $15$ pixels to be the threshold for considering the image alignment viable for correction ($t_f$). The threshold is computed based on the dominant value of optical flow magnitude for the given triplet (considering optical flows with respect to the reference frame - see Figure \ref{fig:md_alg_histo}). The motivation for choosing a threshold value is to eliminate big movements of camera, for which estimating the homography transformation may lead to imprecise HDR fusion. To provide more insight, we report histograms of dominant value of optical flow magnitude for datasets used in our experiments - see Figure \ref{fig:threshold_hist}.
We observe close to unimodal distribution for Chen \cite{Chen21} datasets suggesting that they were collected in \textit{pseudo-static} manner. With such a distribution it is desirable to correct the alignment of most of the LDR samples. On the other hand, Kalantari \cite{Kalantari17} dataset characterises with close to bimodal distribution of camera movements with a strong peak in close to $0$ values and a small increase in values from $20$ to $40$. In this case, we want to set the threshold to categorise most of the second mode as not suitable for alignment. Therefore, we have chosen to consider $15$ pixels as a good threshold that separates modes of Kalantari distribution, whereas preserves almost all Chen LDR images.

Similar reasoning was used in choosing the amount of movement augmentation - drawn from the following Gaussian distributions: \textit{pseudo-static} - $\mathcal{N}(0, 4)$, bigger movements: $\pm\mathcal{N}(20, 3)$. Note that \textit{pseudo-static} augmentation ensures the augmentation within alignment threshold and corresponds to the main mode of movement magnitudes across the images (Fig. \ref{fig:threshold_hist}). Values corresponding to bigger movement lie outside the alignment threshold and try to account for patches that may be discarded in the process of generation within motion domain.

\subsection{Generalisability}

We note that the choice of hyper-parameters is not fine-tuned to any particular dataset but rather inspired by intuition from the data collection process and previous literature. 
For the motion domain we consider optical flow threshold similar to values observed for a static, hand-held camera. 
In the exposure domain the range of illumination for well-exposed regions is inspired by the triangular function weights used in early model-based approaches whose underlying principle is an increasing confidence for pixel values closer to 0.5 [6, 14].
Our hyper-parameters are not fine-tuned but rather fixed for all sub-datasets (Kalantari, D, SRM, HdM2, DnGT, S) used in the paper, hinting robustness and little data-dependency.


\section{Future Work}
In this section we extend further the main manuscript's discussion with respect to limitations of our proposed approach and possible future work avenues which we could not fit within the page limit.

\textbf{Camera Response Function}: In our approach we assume that a gamma transformation can approximate accurately enough the Camera Response Function (CRF) necessary to bring images into \textit{exposure alignment} in the linear domain. This holds true for the most commonly used multi-frame HDR datasets included in the paper (the Kalantari \etal~\cite{Kalantari17} and Chen \etal~\cite{Chen21} datasets).
For the case where CRF is arbitrarily complex and unknown, we would need to estimate the CRF, similarly to \eg \citeappendix{Liu20}.

\textbf{Noise Modelling}: We note that carefully collected dataset presented by Chen \etal~\cite{Chen21} and Kalantari \etal~\cite{Kalantari17} contain images charaterised with negligible amount of noise. Therefore, we do not include a noise model in the process of patch classification and LDR modelling. When working with highly noisy data, patch classification process would have to be adjusted to match the noise level between generated pseudo-labels and source images. Additionally, LDR modelling process would account for noise parameters as in \cite{Hasinoff10}.

\textbf{Motion Modelling}: In our work we introduce the use of motion augmentation in order to further increase and balance the number of training examples where there is misalignment among input LDR frames, as this is crucial to learn LDR-to-HDR mappings for dynamic scenes. We use for this purpose translation-only motion, as despite its simplicity it has proven to be effective. Other more sophisticated transformations could be explored, such as a perspective transform (where the transformation parameters are fitted to represent the target motion distribution), or motion transfer as recently done in \cite{Prabhakar21}.

\textbf{Weak supervision and pre-training}: In our manuscript we explore the full self-supervision scenario as we believe this is the most novel and interesting aspect of our work, however we argue that our approach could be easily used as a self-supervised "pre-training" step that can be later on fine-tuned on available annotated data (\ie weakly supervised), and will explore in the future the impact of size on both unlabelled and labelled datasets to have a deeper understanding of the dynamics between weakly and self-supervised set-ups within our proposed strategy.

\clearpage

\bibliographystyle{splncs04}
\bibliography{egbib}
\bibliographystyleappendix{splncs04}
\bibliographyappendix{egbib}
\end{document}